\documentclass{article}

\usepackage{microtype}
\usepackage{graphicx}
\usepackage{booktabs} %
\usepackage{caption}
\usepackage{subcaption}
\usepackage[noend]{algpseudocode}
\usepackage{enumitem}
\usepackage{amsmath}
\usepackage{amssymb}

\usepackage{hyperref}

\usepackage{mathtools}

\DeclarePairedDelimiterX{\inner}[2]{\langle}{\rangle}{#1, #2}

\usepackage{multirow}

\usepackage[accepted]{icml2021}

\renewcommand{\v}[1]{\ensuremath{\boldsymbol{#1}}}

\newcommand{\x}{\ensuremath{\v{x}}}

\newcommand{\q}{\theta}

\newcommand{\KL}[2]{\ensuremath{\mathrm{KL}\left({#1} \:\middle\vert\middle\vert\: {#2}\right)}}

\newcommand{\hide}[1]{}

\makeatletter
\DeclareRobustCommand{\cev}[1]{%
  \mathpalette\do@cev{#1}%
}
\newcommand{\do@cev}[2]{%
  \fix@cev{#1}{+}%
  \reflectbox{$\m@th#1\vec{\reflectbox{$\fix@cev{#1}{-}\m@th#1#2\fix@cev{#1}{+}$}}$}%
  \fix@cev{#1}{-}%
}
\newcommand{\fix@cev}[2]{%
  \ifx#1\displaystyle
    \mkern#23mu
  \else
    \ifx#1\textstyle
      \mkern#23mu
    \else
      \ifx#1\scriptstyle
        \mkern#22mu
      \else
        \mkern#22mu
      \fi
    \fi
  \fi
}

\icmltitlerunning{Conjugate Energy-Based Models}

\begin{document}

\twocolumn[
\icmltitle{Conjugate Energy-Based Models}

\icmlsetsymbol{equal}{*}

\begin{icmlauthorlist}
\icmlauthor{Hao Wu}{equal,neu}
\icmlauthor{Babak Esmaeili}{equal,neu}
\icmlauthor{Michael Wick}{oracle}
\icmlauthor{Jean-Baptiste Tristan}{bc}
\icmlauthor{Jan-Willem van de Meent}{neu}
\end{icmlauthorlist}

\icmlaffiliation{neu}{Khoury College of Computer Sciences, Northeastern University, Boston, MA, USA}
\icmlaffiliation{oracle}{Oracle Labs, MA, USA}
\icmlaffiliation{bc}{Computer Science department, Boston College, MA, USA}

\icmlcorrespondingauthor{Hao Wu}{wu.hao10@northeastern.edu}
\icmlcorrespondingauthor{Babak Esmaeili}{esmaeili.b@northeastern.edu}

\icmlkeywords{Machine Learning, ICML}

\vskip 0.3in
]

\printAffiliationsAndNotice{\icmlEqualContribution} %

\begin{abstract}
In this paper, we propose conjugate energy-based models (CEBMs), a new class of energy-based models that define a joint density over data and latent variables. The joint density of a CEBM decomposes into an intractable distribution over data and a tractable posterior over latent variables. CEBMs have similar use cases as variational autoencoders, in the sense that they learn an unsupervised mapping from data to latent variables. However, these models omit a generator network, which allows them to learn more flexible notions of similarity between data points. Our experiments demonstrate that conjugate EBMs achieve competitive results in terms of image modelling, predictive power of latent space, and out-of-domain detection on a variety of datasets. 
\end{abstract}

\vspace{-1.5\baselineskip}
\section{Introduction}
\label{sec:intro}

Deep generative models approximate a data distribution by combining a prior over latent variables with a neural generator, which maps latent variables to points on a data manifold. It is common to evaluate these models in terms of their ability to generate realistic examples, or their estimated densities for unseen data. However, an arguably more important use case for these models is unsupervised representation learning. If a generator can faithfully represent the data in terms of a lower-dimensional set of latent variables, then we hope that these variables will encode a set of semantically meaningful factors of variation that will be relevant to a broad range of downstream tasks.

Guiding a model towards a semantically meaningful representation requires some form of inductive bias. A large body of work on variational autoencoders (VAEs, \cite{kingma2013auto-encoding,rezende2014stochastic}) has explored the use of priors as inductive biases. Relatively mild biases in the form of conditional independence are common in the literature on disentangled representations \citep{higgins2016beta,kim2018disentangling,chen2018isolating,esmaeili2019structured}. More generally, recent work has shown that defining priors that reflect the structure of the underlying data will lead to representations that are easier to interpret and generalize better. Examples include priors that represent objects in an image \citep{eslami2016air,lin2020space,engelcke2019genesis,crawford2019spatially}, or moving objects in video \citep{crawford2019exploiting, kosiorek2018sequential,wu2020amortized,lin2020improving}.

Despite steady progress, work on disentangled representations and structured VAEs still predominantly considers synthetic data. VAEs employ a neural generator that is optimized to reconstruct examples in the training set. For complex natural scenes, learning a generator that can produce pixel-perfect reconstructions poses fundamental challenges, given the combinatorial explosion of possible inputs. This is not only a problem for generation, but also from the perspective of the learned representation; a VAE must encode all factors of variation that give rise to large deviations in pixel space, regardless of whether these factors are semantically meaningful (e.g.~presence and locations of objects) or not (e.g.~shadows of objects in the background of the image).

\begin{figure*}[!t]
\centering
\includegraphics[width=\textwidth]{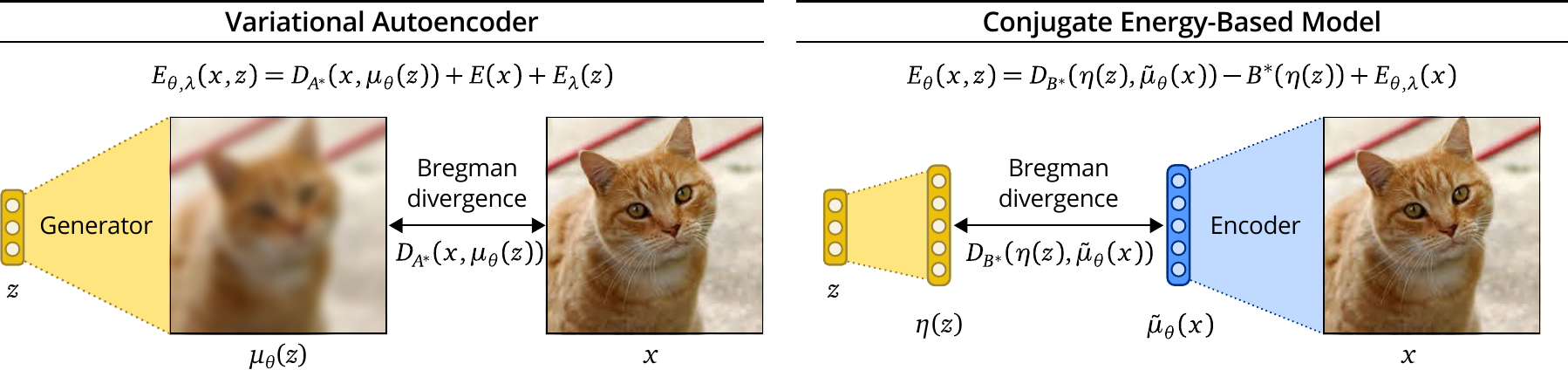}
\vspace*{-4.5ex}
\caption{Comparison between a VAE and a CEBM. A variational autoencoder with a Gaussian or Bernoulli likelihood has an energy that can be expressed in terms of a Bregman divergence in the data space $D_{A^*}(x, \mu_\theta(z))$ between an image $x$ and the reconstruction from the generator network $\mu_\theta(z)$. The energy function in a CEBM can be expressed in terms of a Bregman divergence in the latent space $D_{B^*}(\eta(z), \tilde{\mu}_\theta(x))$ between a vector of natural parameters $\eta(z)$ and the output of an encoder network $\tilde{\mu}_\theta(x)$. See main text for details.}
\vspace*{-0.0ex}
\label{fig:overview}
\end{figure*}

The motivating question that we consider in this paper is whether it is possible to train latent-variable models without minimizing  pixel-level discrepancies between an image and its reconstruction. Instead, we would like to design an objective that minimizes the discrepancy between the encoding of an image and the latent variables, which will in general be in a lower-dimensional space compared to the input. Our hope is that doing so will allow a model to learn more abstract representations, in the sense that it becomes easier to discard factors of variation that give rise to variation in pixel space, but should be considered noise.

In this paper, we consider energy-based models (EBMs) with latent variables as a particular instantiation of this general idea. %
EBMs with latent variables are by no means new; they have a long history in the context of restricted Boltzmann machines (RBMs) and related models \citep{smolensky1986information,hinton2002training,welling2005exponential}. 
Our motivation in the present work is to design a class of EBMs that retain the desirable features of VAEs, but employ a discriminative energy function to model data at an intermediate level of representation that does not necessarily encode all features of an image at the pixel level.

Concretely, we propose conjugate EBMs (CEBMs), a new family of energy-based latent-variable models in which the energy function defines a neural exponential family. While the normalizer of CEBMs is intractable, we can nonetheless compute the posterior in closed form when we pair the likelihood with an appropriate conjugate bias term. As a result, the neural sufficient statistics in a CEBM fully determine both the marginal likelihood and the encoder, thereby side-stepping the need for a generator (Figure~\ref{fig:overview}).

Our contributions can be summarized as follows:
\begin{enumerate}[noitemsep,topsep=0pt,parsep=6pt,partopsep=0pt]
    \item We propose CEBMs, a class of energy-based models for unsupervised representation learning. The density of a CEBM factorizes into a tractable posterior and an energy-based marginal over data. This means that CEBMs can be trained using existing methods for EBMs, whilst inference is tractable at test time. 
    \item Unlike VAEs, CEBMs model data not at the pixel level, but at the level of the latent representation. We interpret the energy function of CEBMs in terms of a Bregman divergence in the latent space, and show that the density of a VAE can similarly be expressed in terms of a Bregman divergence in the data space. 
    \item We show that two of the most common inductive biases in VAEs can be incorporated in CEBMs: a spherical Gaussian and a mixture of Gaussians. 
    \item We evaluate how well CEBMs learned representations agree with class labels (which are not used during training). We show that neighbors are more likely to belong to the same class, which translates to increased performance in downstream classification tasks. Moreover, CEBMs perform competitively in out-of-domain detection. We do also note limitations; in particular we observe that CEBMs suffer from posterior collapse.
\end{enumerate}

\vspace*{-0.75ex}
\section{Background}
\label{sec:background}
\vspace*{-0.25ex}

\subsection{Energy-Based Models}
An EBM~\citep{lecun2006tutorial} defines a probability density for $x\in\mathbb{R}^D$ via the Gibbs-Boltzmann distribution
\begin{align*}
    p_\q(x) &= \frac{\exp \left\{- E_\q(x)\right\}}{Z_\q},
    &
    Z_\q &= \int \!dx \: \exp \{- E_\q(x)\}.
\end{align*}
The function $E_\q : \mathbb{R}^D \xrightarrow[]{} \mathbb{R}$ is called the energy function which maps each configuration to a scalar value, the energy of the configuration. This type of model is widely used in statistical physics, for example in Ising models.
The distribution can only be evaluated up to an unknown constant of proportionality, since computing the normalizing constant $Z_\q$ (also known as the partition function) requires an intractable integral with respect to all possible inputs $x$.

Our goal is to learn a model $p_\q(x)$ that is close to the true data distribution $p_\text{data}(x)$. A common strategy is to minimize the Kullback-Leibler divergence between the data distribution and the model, which is equivalent to maximizing the expected log-likelihood
\begin{align}
\label{eq:obj-ebm}
\mathcal{L}(\q)
&= \mathbb{E}_{p_\text{data}(x)}[\log p_\q (x)]
,\\ \nonumber
&= \mathbb{E}_{p_\text{data}(x)}[-E_\q (x)] - \log Z_\q.
\end{align}
The key difficulty when performing maximum likelihood estimation is that computing the gradient of $\log Z_\q$ is intractable. This gradient can be expressed as an expectation with respect to $p_\q(x)$,
\begin{align}
    \nabla \log Z_\q 
    &= 
    \mathbb{E}_{p_\q(x')}
    \left[
    -\nabla_\q E_\q(x')
    \right]
    ,
\end{align}
which means that the gradient of $\mathcal{L}(\q)$ has the form:
\begin{align*}
\nabla_\q \mathcal{L}(\q)
&=
- \mathbb{E}_{p_\text{data}(x)}[\nabla_\q E_\q (x)] + \mathbb{E}_{p_\q(x')}[\nabla_\q E_\q(x')]
.
\end{align*}
This corresponds to \emph{maximizing} the probability of samples $x \sim p_\text{data}(x)$ from the data distribution and \emph{minimizing} the probability of samples $x' \sim p_\q(x')$ from the learned model. 

Contrastive divergence methods \cite{hinton2002training} compute a Monte Carlo estimate of this gradient, which requires a method for approximate inference to generate samples $x' \sim p_\q(x')$. A common method for generating samples from EBMs is Stochastic Gradient Langevin Dynamics (SGLD,~\cite{welling2011bayesian}), which initializes a sample $x'_0 \sim p_0(x')$ and performs a sequence of gradient updates with additional injected noise $\epsilon$,
\begin{align}
\label{eq:sgld}
x'_{i+1} &= x'_i - \frac{\alpha}{2} \frac{\partial E_\q (x')}{\partial x'} + \epsilon
\,,&
\epsilon &\sim N(0, \alpha)
.
\end{align}
SGLD is motivated as a discretization of a stochastic differential equation whose stationary distribution is equal to the target distribution.
It is correct in the limit $i \to \infty$ and $\alpha \to 0$, but in practice will have a bias. 

The initialization $x'_0$ is crucial because it determines the number of steps needed to converge to a high-quality sample. For this reason, EBMs are commonly trained using persistent contrastive divergence (PCD,~\cite{du2019implicit,tieleman2008training}), which initializes some samples from a replay buffer $\mathcal{B}$ of previously generated samples~\cite{nijkamp2019anatomy, du2019implicit, xie2016theory}.

\subsection{Energy-Based Latent-Variable Models}

Energy-based latent-variable models are a subclass of EBMs where the the energy function defines joint density on observed data $x \in \mathbb{R}^D$ and latent variable $z \in \mathbb{R}^K$,
\begin{equation}
    p_\q(x,z) = \frac{\exp \left\{- E_\q(x,z)\right\}}{Z_\q}.
\end{equation}
Some of the most well-known examples of this family of models include restricted Boltzmann machines (RBMs,~\citep{smolensky1986information,hinton2002training}), deep belief nets (DBNs,~\cite{hinton2006fast}), and deep Boltzmann machines (DBMs,~\citep{salakhutdinov2009deep}). 

Similar to standard EBMs, energy-based latent-variable models can also be trained using contrastive divergence methods, where the gradient of $\mathcal{L}(\q)$ can be expressed as:
\begin{align*}
- \mathbb{E}_{p_\text{data}(x)p_\q(z|x)}[\nabla_\q E_\q (x,z)] + \mathbb{E}_{p_\q(x',z')}[\nabla_\q E_\q(x',z')]
.
\end{align*}
Estimating this gradient has the additional problem of requiring samples from the posterior $p_\q(z|x)$ which is also intractable in general. %

\subsection{Conjugate Exponential Families}
An exponential family is a set of distributions whose probability density can be expressed in the form
\begin{align}
\label{eq:exponential-family}
    p(x \mid \eta) 
    &= 
    h(x) \exp \big\{ 
        \inner{t(x)}{\eta}    
        - A(\eta) \big\},
\end{align}
where $h: \mathcal{X} \to \mathbb{R}^+$ is a base measure, $\eta \in \mathcal{H} \subseteq \mathbb{R}^K$ is a vector of natural parameters, $t: \mathcal{X} \to \mathbb{R}^K$ is a vector of sufficient statistics, and $A: \mathcal{H} \to \mathbb{R}$ is the log normalizer (or cumulant function),
\begin{align}
    \label{eq:log-normalizer}
    A(\eta) = \log Z(\eta) = \int \!\! dx \: h(x) \exp \big\{ \inner{t(x)}{\eta}  \big\}.
\end{align}
If a likelihood belongs to an exponential family, then there exists a conjugate prior that is itself an exponential family
\begin{align}
    p(\eta \mid \lambda, \nu) 
    &= 
    \exp \big\{ 
        \inner{\eta}{\lambda} - A(\eta) \nu - B(\lambda, \nu)
    \big\}.
\end{align}   
The convenient property of conjugate exponential families is that both the marginal likelihood $p(x \mid \lambda, \nu)$ and the posterior $p(\eta \mid x, \lambda, \nu)$ are tractable. If we define 
\begin{align}
  \label{eq:cef-posterior-params}
  \tilde{\lambda}(x) = \lambda + t(x), 
  \qquad
  \tilde{\nu}= \nu + 1,
\end{align}
then the posterior and marginal likelihood are
\begin{align}
    \label{eq:cef-posterior-and-marginal}
    \begin{split}
    p(\eta \mid x, \lambda, \nu) &= p(\eta \mid \tilde{\lambda}(x), \tilde{\nu}),
    \\
    p(x \mid \lambda, \nu) &= h(x) \: \exp\big\{ B(\tilde{\lambda}(x), \tilde{\nu}) - B(\lambda, \nu) \big\}.
    \end{split}
\end{align}
\subsection{Legendre Duality in Exponential Families}
Two convex functions $A: \mathcal{H} \to \mathbb{R}^+$ and $A^*: \mathcal{M} \to \mathbb{R}^+$ on spaces $\mathcal{H} \subseteq \mathbb{R}^K$ and $\mathcal{M} \subseteq \mathbb{R}^K$ are conjugate duals when
\begin{align}
    A^*(\mu) := \sup_{\eta \in \mathcal{H}} \big\{ \inner{\mu}{\eta} - A(\eta) \big\}.
\end{align}
When $A$ is a function of Legendre type (see \citet{rockafellar1970convex} for details), the gradients of these functions define a bijection between conjugate spaces by mapping points to their corresponding suprema
\begin{align}
    \eta(\mu) &= \nabla A^*(\mu),
    &
    \mu(\eta) &= \nabla A(\eta), 
\end{align}
such that we can express $A^*(\mu)$ at the supremum as
\begin{align}
    A^*(\mu) &= \inner{\mu}{\eta(\mu)} - A(\eta(\mu))
\end{align}
The log normalizer $A(\eta)$ of an exponential family is of Legendre type when the family is regular and minimal ($\mathcal{H}$ is an open set and sufficient statistics $t(x)$ are linearly independent; see \citet{wainwright2008graphical} for details). We refer to $\mathcal{M}$ as the mean parameter space, since we can express any $\mu \in \mathcal{M}$ as the expected value of the sufficient statistics
\begin{align}
    \mu(\eta) = \mathbb{E}_{p(x \mid \eta)}[t(x)].
\end{align}

\subsection{Bregman Divergences and Exponential Families}

A Bregman divergence for a function $F: \mathcal{M} \to \mathbb{R}$ that is continuously-differentiable and strictly convex on a closed set $\mathcal{M}$ has the form 
\begin{align}
    \begin{split}
    D_F(\mu', \mu) &= F(\mu') - F(\mu) 
    - \inner{\mu'-\mu}{\nabla F(\mu)}
    \end{split}
    .
\end{align}
Well-known special cases of Bregman divergences include the squared distance ($F(\mu) = \inner{\mu}{\mu}$) and the Kullback-Leiber (KL) divergence ($F(\mu) = \sum_{k} \mu_k \log \mu_k$). 

Any Bregman divergence can be associated with an exponential family and vice versa, where $F(\mu)=A^*(\mu)$ is the conjugate dual of $A(\eta)$ (see \citet{banerjee2005clustering}). To see this, we re-express the log density of a (regular and minimal) exponential family using the substitution $\mu = \nabla A(\eta)$\footnote{We here omit the base measure $h(x)$ for notational simplicity.},
\begin{align}
    \label{eq:bregman-density}
    \begin{split}
    \log p(x \mid \eta) &= \inner{t(x)}{\eta} - A(\eta), \\
                        &= \big(\inner{\mu}{\eta} -A(\eta)  \big)  + \inner{t(x)\!-\!\mu}{\eta},\\
                        &= A^*(\mu) + \inner{t(x)\!-\!\mu}{\nabla A^*(\mu)}, \\
                        &= - D_{A^*}(t(x), \mu) + A^*(t(x)).
    \end{split}
\end{align}
In other words, the log density of an exponential family can be expressed in terms of a bias term  $A^*(t(x))$\footnote{Or $A^*(t(x)) + \log h(x)$ when we include $h(x)$ the density.}, and a notion of agreement in the form of a Bregman divergence $D_{A^*}(t(x), \mu)$ between the sufficient statistics $t(x)$ and the mean parameters $\mu$. We will make use of this property of exponential families to provide an interpretation of both CEBMs and VAEs in terms of Bregman divergences.

\section{Conjugate Energy-Based Models}
\label{sec:cebm}
\vspace*{-0.25ex}

We are interested in learning a probabilistic model that defines a joint density $p_{\q,\lambda}(x, z)$ over high-dimensional data $x \in \mathbb{R}^D$ and a lower-dimensional set of latent variables $z \in \mathbb{R}^K$. The intuition that guides our work is that we would like to measure agreement between latent variables and data at a high level of representation, rather than at the level of individual pixels, where it may be more difficult to distinguish informative features from noise. To this end, we will explore energy-based models as an alternative to VAEs. 

Concretely, we propose to consider models of the form
\begin{align}
    p_{\q,\lambda}(x, z) &= \frac{1}{Z_{\q,\lambda}}\: \exp \big\{ -E_{\q,\lambda}(x,z)\big\},
\end{align}
where the energy function takes a form that is inspired by exponential family distributions
\begin{align}
    \label{eq:cebm-energy}
    E_{\q,\lambda}(x,z) = -\inner{t_\q(x)}{\eta(z)} + E_{\lambda}(z).
\end{align}
In this energy function, $\theta$ are the weights of a network $t_\q: \mathbb{R}^D \to \mathbb{R}^H$, which plays the role of an encoder by mapping high-dimensional data to a lower-dimensional vector of neural sufficient statistics. The function $\eta: \mathbb{R}^K \to \mathbb{R}^H$ maps latent variables to a vector of natural parameters in the same space as the neural sufficient statistics. The function $E_\lambda: \mathbb{R}^K \to \mathbb{R}$ serves as an inductive bias, with hyperparameters $\lambda$, that plays a role analogous to the prior.

We will consider a bias $E_\lambda(z)$ in form of a tractable exponential family with sufficient statistics $\eta(z)$ 
\begin{align}
    E_\lambda(z) = -\log p_\lambda(z) =  -\inner{\eta(z)}{\lambda} + B(\lambda).
\end{align}
We can then express the energy function as
\begin{align}
    E_{\q,\lambda}(x, z) = -\inner{\lambda + t_\q(x)}{\eta(z)} + B(\lambda).
\end{align}
This form of the energy function has a convenient property: It corresponds to a model $p_{\theta,\lambda}(x,z)$ in which the posterior $p_{\theta,\lambda}(z \mid x)$ is tractable. To see this, we make a substitution $\tilde{\lambda}_\theta(x) = \lambda + t_\q(x)$ analogous to the one in Equation~\ref{eq:cef-posterior-params}, which allows us to express the energy  as
\begin{align}
    E_{\q,\lambda}(x, z)
    &\!=
    \!-\inner{\eta(z)}{\tilde{\lambda}_\theta(x)} 
    \!+\! B(\tilde{\lambda}_\theta(x)) 
    \!+\! E_{\theta,\lambda}(x), \\
    E_{\theta,\lambda}(x) 
    &=
    - B(\tilde{\lambda}_\theta(x)) 
    + B(\lambda) .
\end{align}
We see that we can factorize the corresponding density
\begin{align}
    p_{\q,\lambda}(x,z) 
    &= 
    p_{\q,\lambda}(x) \: p_{\q,\lambda}(z \mid x), 
\end{align}
which yields a posterior and marginal that are analogous the distributions in Equation~\ref{eq:cef-posterior-and-marginal}%
\begin{align}
    p_{\q,\lambda}(z \mid x) &= p(z \mid \tilde{\lambda}_\q(x)),\\
    \begin{split}
    p_{\q,\lambda}(x) 
    &= \frac{1}{Z_{\q,\lambda}} \exp\big\{ -E_{\theta,\lambda}(x) \}, \\
    &= \frac{1}{Z_{\q, \lambda}} \exp\big\{ B\big(\tilde{\lambda}_\q(x) \big) - B\big(\lambda\big) \big\}.
    \end{split}
    \label{eq:cebm-marginal}
\end{align}
In other words, the joint density of this model factorizes into a tractable posterior $p_{\theta,\lambda}(z \mid x)$ and an intractable energy-based marginal likelihood $p_{\theta,\lambda}(x)$. This posterior is conjugate, in the sense that it is in the same exponential family as the bias. For this reason, we refer to this class of models as conjugate energy-based models (CEBMs).

\vspace*{-1.0ex}
\section{Relationship to VAEs}
\vspace*{-0.25ex}

CEBMs differ from VAEs in that they lack a generator network. Instead, the density is fully specified by the encoder network $t_\theta(x)$, which defines a notion of agreement $\inner{\tilde{\lambda}_\theta(x)}{\eta(z)}$ between data and latent variables in the latent space. As with other exponential families, we can make this notion of agreement explicit by expressing the conjugate posterior in terms of a Bregman divergence using the decomposition in Equation~\ref{eq:bregman-density}
\begin{align}
    \begin{split}
    E_{\theta,\lambda}(x, z) 
    &=
    D_{B*}(\eta(z), \tilde{\mu}_\theta(x)) \\
    &\qquad
    - B^*(\eta(z)) + E_{\theta,\lambda}(x).
    \end{split}
\end{align}
Here $B^*(\mu)$ is the conjugate dual of the the log normalizer $B(\lambda)$, and we use $\tilde{\mu}_\theta(x) = \mu(\tilde{\lambda}_\theta(x))$ as a shorthand for the mean-space posterior parameters. We see that maximizing the density corresponds to minimizing a Bregman divergence in the space of sufficient statistics of the bias. 

In Figure~\ref{fig:overview}, we compare CEBMs to VAE in terms of the energy function for the log density of the generative model. In making this comparison, we have to keep in mind that these models are trained using different methods, and that VAEs have a tractable density $p_\theta(x,z)$. That said, the objectives in both models maximize the marginal likelihood, so we believe that it is instructive to write down the corresponding Bregman divergence in the VAE likelihood. This likelihood is typically a Gaussian with known variance, or a Bernoulli distribution (when modeling binarized images). Both distributions have sufficient statistics $t(x)=x$. Once again omitting the base measure $h(x)$ for expediency, we can express the log density of a VAE as an energy
\begin{align}
\begin{split}
&E_{\theta,\lambda}(x,z) = -\log p_\theta(x | z) - \log p_\lambda(z), \\
&\qquad 
= -\inner{x}{\eta_\theta(z)} + A(\eta_\theta(z)) - \log p_\lambda(z).\\
&\qquad 
= D_{A^*}(x, \mu_\theta(z)) - A^*(x) - \log p_\lambda(z)
\end{split}
\end{align}
Here $A^*(x)$ is the conjugate dual of the log normalizer $A(\eta)$, and we use $\eta_\theta(z)$ and $\mu_\theta(z)$ to refer to the output of the generator network in the natural-parameter and the mean-parameter space respectively. To reduce clutter and accommodate the case where a base measure $h(x)$ is needed (e.g.~that of a Gaussian likelihood with known variance), we will introduce the additional shorthands
\begin{align}
    \label{eq:vae-bias}
    E(x) &= - A(x) \!-\! \log h(x),
    &
    E_\lambda(z) &= - \log p_\lambda(z).
\end{align}
We then see that the energy function of a VAE has the form 
\begin{align}
    E_{\theta,\lambda}(x,z) 
    &=
    D_{A^*}(x, \mu_\theta(z))
    +
    E(x)
    + 
    E_\lambda(z).
\end{align}
Like that of a CEBM, the energy function of a VAE contains a Bregman divergence, as well as two terms that depend only on $x$ and $z$. However, whereas the Bregman divergence in CEBM is defined in the mean-parameter space of the latent variables, that of a VAE is computed in the data space.

\vspace*{-0.75ex}
\section{Inductive Biases}
\label{sect:inductive-biases}
\vspace*{-0.5ex}
CEBMs have a property that is somewhat counter-intuitive. While the posterior $p_{\q,\lambda}(z \mid x)$ in this class of models is tractable, the prior is in general not tractable. In particular, although the bias $-E_\lambda(z)$ is the logarithm of a tractable exponential family, it is not the case that $p_{\q,\lambda}(z) = p_{\lambda}(z)$. Rather the prior $p_{\q,\lambda}(z)$ has the form,
\begin{align*}
    p_{\q,\lambda}(z) = \frac{\exp \{-E_\lambda(z)\}}{Z_{\q,\lambda}} \ \int \! dx \: \exp \{\inner{t_\q(x)}{\eta(z)}\}.
\end{align*}
In other words, $E_\lambda(z)$ defines an inductive bias, but this bias is different from the tractable prior in a VAE\footnote{The bias in a VAE contains the log prior $\log p_\lambda(z)$ and the log normalizer $A(\eta_\q(z))$ of the likelihood. In a CEBM, by contrast, we omit the term $A_\q(\eta(z)) = \log \int dx \exp \{ \inner{t_\q(x)}{\eta(z)} \}$, which is intractable, and hereby implicitly absorb it into its prior.}, in the sense that it imposes only a soft constraint on the geometry of the latent space.

In principle, the bias in a CEBM can take the form of any exponential family distribution. Since products of exponential families are also in the exponential family, this covers a broad range of possible biases. For purposes of evaluation in this paper, we will constrain ourselves to two cases:  

\vspace*{-0.2\baselineskip}
\paragraph{1. Spherical Gaussian.} As a bias that is analogous to the standard prior in VAEs, we consider a spherical Gaussian with fixed hyperparameters $(\mu,\sigma)=(0,1)$ for each dimension of $z \in \mathbb{R}^K$,
\begin{align*}
    E_\lambda(z) = -\sum_{k} \big( \inner{\eta(z_k)}{\lambda} - B(\lambda) \big).
\end{align*}
Each term has sufficient statistics $\eta(z_k) = (z_k, z_k^2)$, natural parameters $\lambda$, and log normalizer $B(\lambda)$ as
\begin{align*}
  &\lambda = 
  \left(
      \frac{\mu}{\sigma^2},
      -\frac{1}{2\sigma^2}
  \right)
  =
  \left(
      0,
      -\frac{1}{2}
  \right)
  ,
  \\
    &B(\lambda) 
    =
    -\frac{\lambda_{1}^2}{4 \lambda_{2}}
    -
    \frac{1}{2} \log (-2\lambda_{2})
    .
\end{align*}
The marginal likelihood of the CEBM is then
\begin{align*}
    p_{\q,\lambda}(x) 
    =
    \frac{1}{Z_{\q,\lambda}}
    \exp \Big\{
      \sum_{k} \big(B(\tilde{\lambda}_{\theta,k}(x)) - B(\lambda)\big)
    \Big\},
\end{align*}
where $\tilde{\lambda}_{\theta,k}(x) = \lambda + t_{\q,k}(x)$ 
and $t_{\q,k}(x)$ is the sufficient statistics that corresponds to $z_k$.

\vspace*{-0.2\baselineskip}
\paragraph{2. Mixture of Gaussians.} In our experiments, we will consider datasets that are normally used for classification. These datasets, by design, exhibit multimodal structure that we would like to see reflected in the learned representation.In order to design a model that is amenable to uncovering this structure, we will extend the energy function in Equation~\ref{eq:cebm-energy} to contain a mixture component $y$
\begin{align*}
    E_{\q,\lambda}(x,y,z) =
        -\inner{t_\q(x)}{\eta(y,z)}
        +
        E_\lambda(y,z).    
\end{align*}
As an inductive bias, we will consider a bias in the form of a mixture of $L$ Gaussians,
\begin{align*}
    E_\lambda(y,z) =
    -\sum_{k,l} I[y=l]
    \big(
        \inner{\eta(z_k)}{\lambda_{l,k}} - B(\lambda_{l,k})
    \big).
\end{align*}
Here $z \in \mathbb{R}^K$ is a vector of features and $y \in \{1, \dots, L\}$ is a categorical assignment variable. The bias for each component $l$ is a spherical Gaussian with hyperparameters $\lambda_{l,k}$ for each dimension $k$. Again, using the notation $\tilde{\lambda}_{\theta,l,k} = \lambda_{l,k} + t_{\q,k}(x)$ to refer to the posterior parameters, then we obtain an energy
\begin{align*}
    E_{\q,\lambda}(x,y,z) =
    -
    \sum_{k,l}
    I[y=l]
    \big(  
        \inner{\eta(z_k)}{\tilde{\lambda}_{\q,l,k}} 
        - B(\lambda_{l,k})
    \big).
\end{align*}
We can then define a joint probability over data $x$ and the assignment $y$ in terms the log normalizer $B(\cdot)$,
\begin{align*}
  & p_{\q,\lambda}(x,y) = \\
  & \qquad  
  \frac{1}{Z_{\q,\lambda}}
  \exp \Big\{ 
    \sum_{k,l} 
    I[y=l]
    \big(
    B(\tilde{\lambda}_{\q,l,k})
    -
    B(\lambda_{l,k})
    \big)
  \Big\},
\end{align*}
which then allows us to compute the marginal $p_{\q,\lambda}(x)$ by summing over $y$. We optimize this marginal with respect hyperaparameters $\lambda_{l,k}$ as well as the weights $\q$.

\begin{figure*}[!t]
\centering
\includegraphics[width=0.235\textwidth]{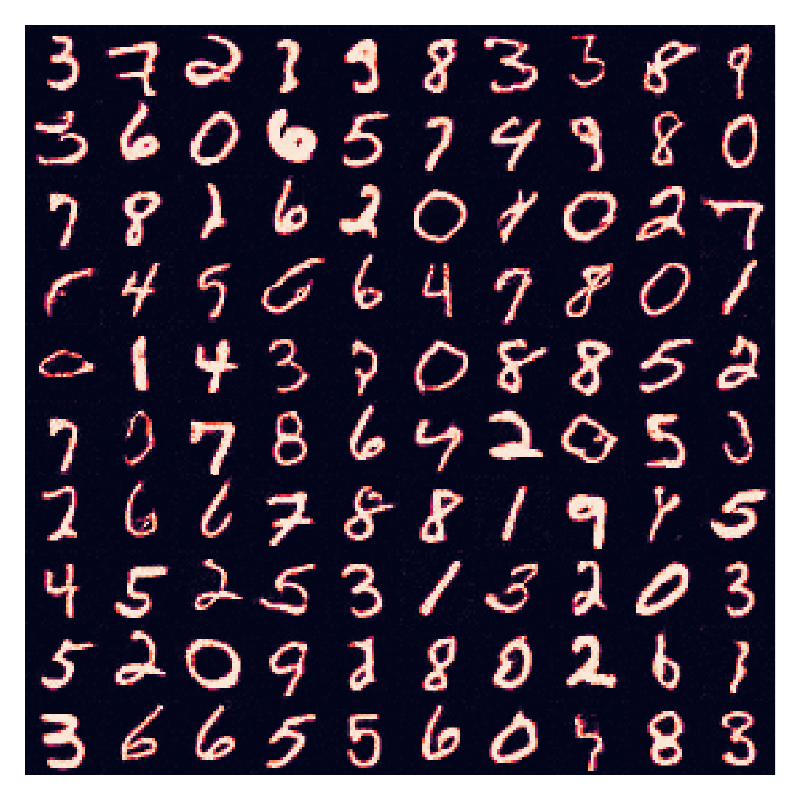}
\includegraphics[width=0.235\textwidth]{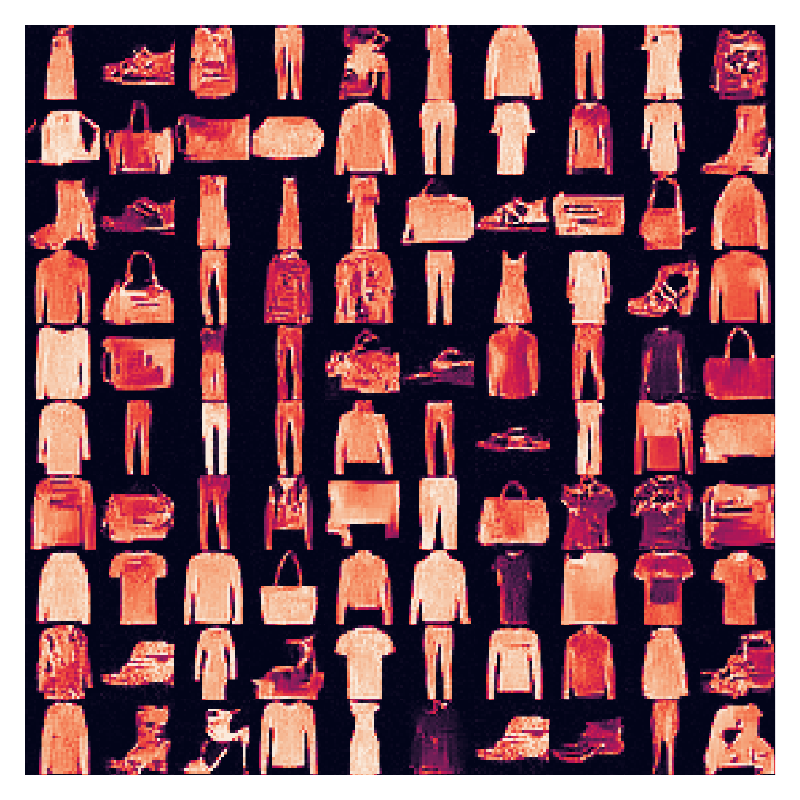}
\includegraphics[width=0.235\textwidth]{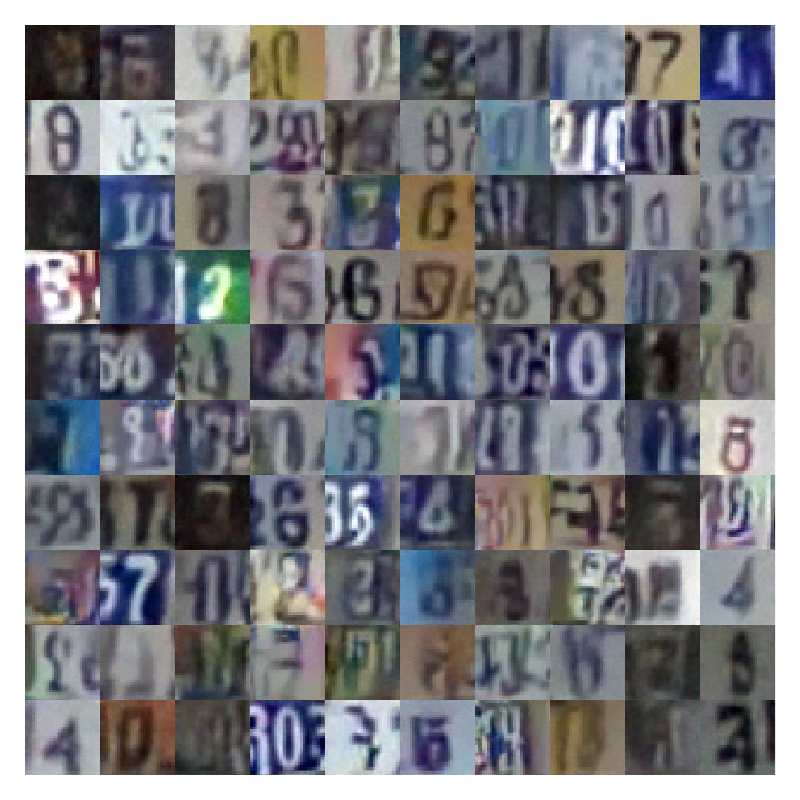}
\includegraphics[width=0.235\textwidth]{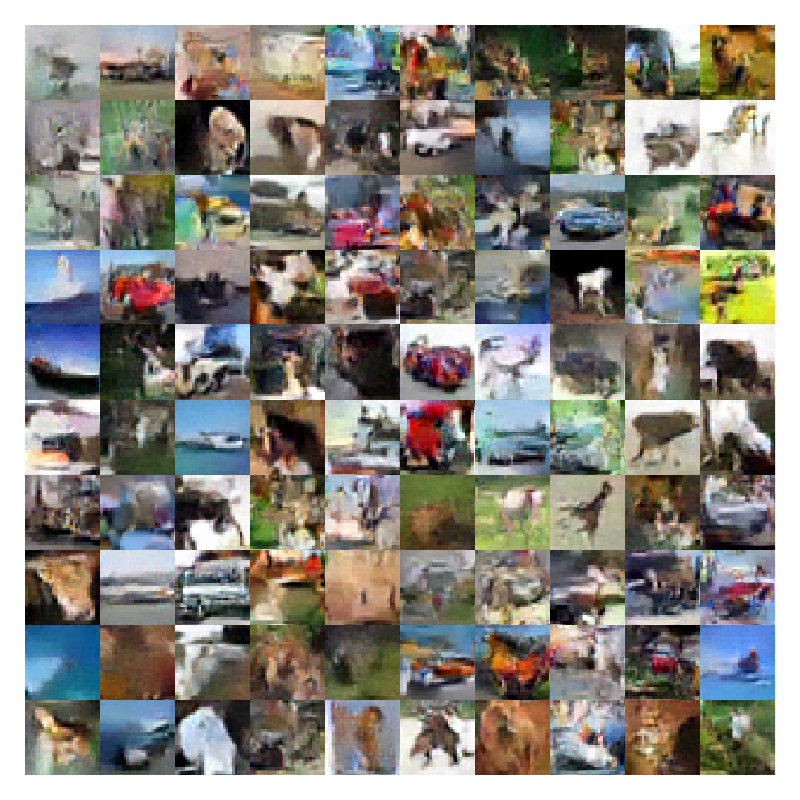}
\vspace*{-2.0ex}
\caption{Samples generated from a CEBM trained on MNIST, Fashion-MNIST, SVHN and CIFAR-10.}
\vspace*{-2.0ex}
\label{fig:generated-samples}
\end{figure*}

\begin{figure*}[!t]
\centering
\begin{subfigure}{0.4\textwidth}
\centering
\includegraphics[width=\linewidth]{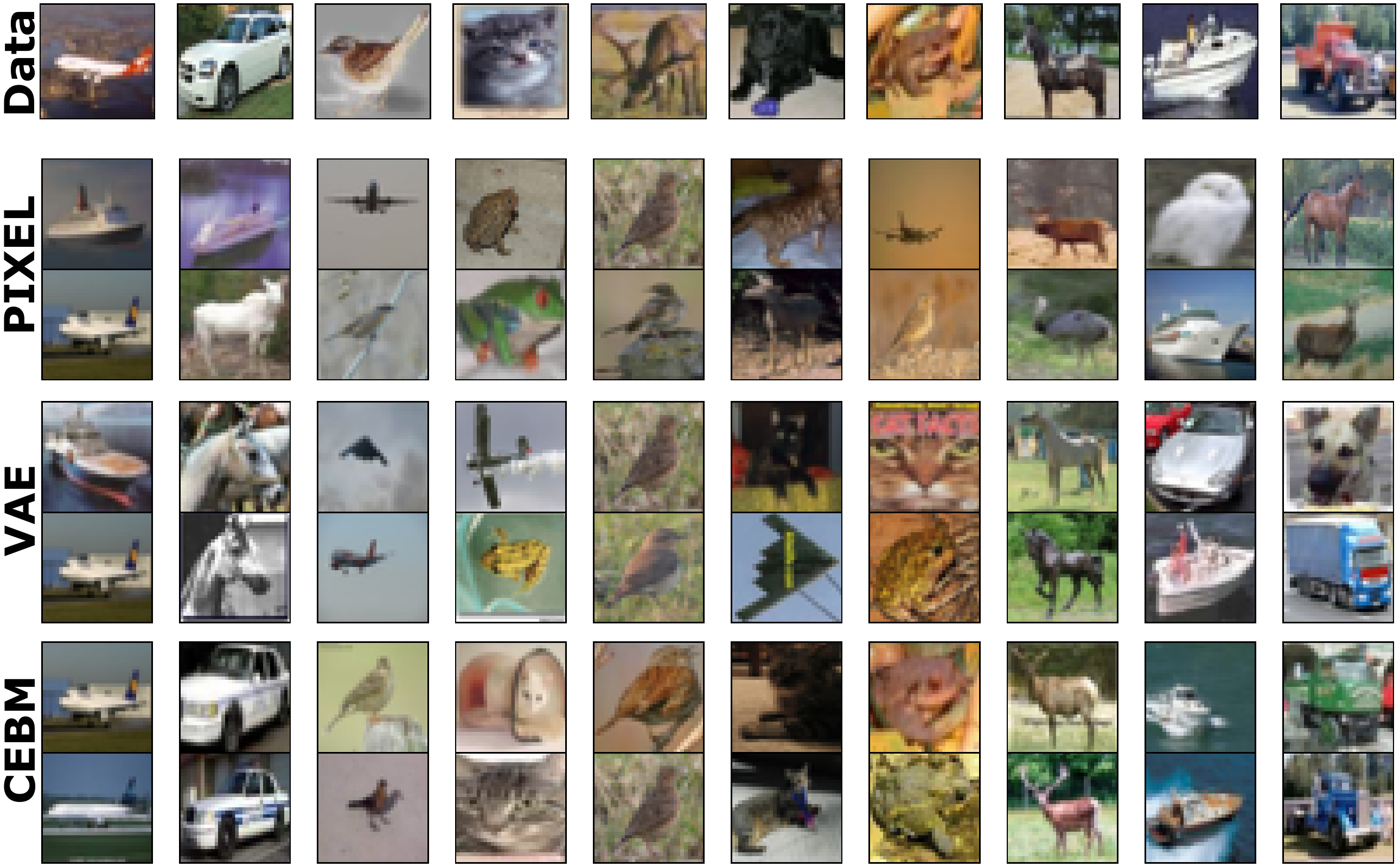}
\end{subfigure}%
\begin{subfigure}{0.55\textwidth}
\centering
\includegraphics[width=\linewidth]{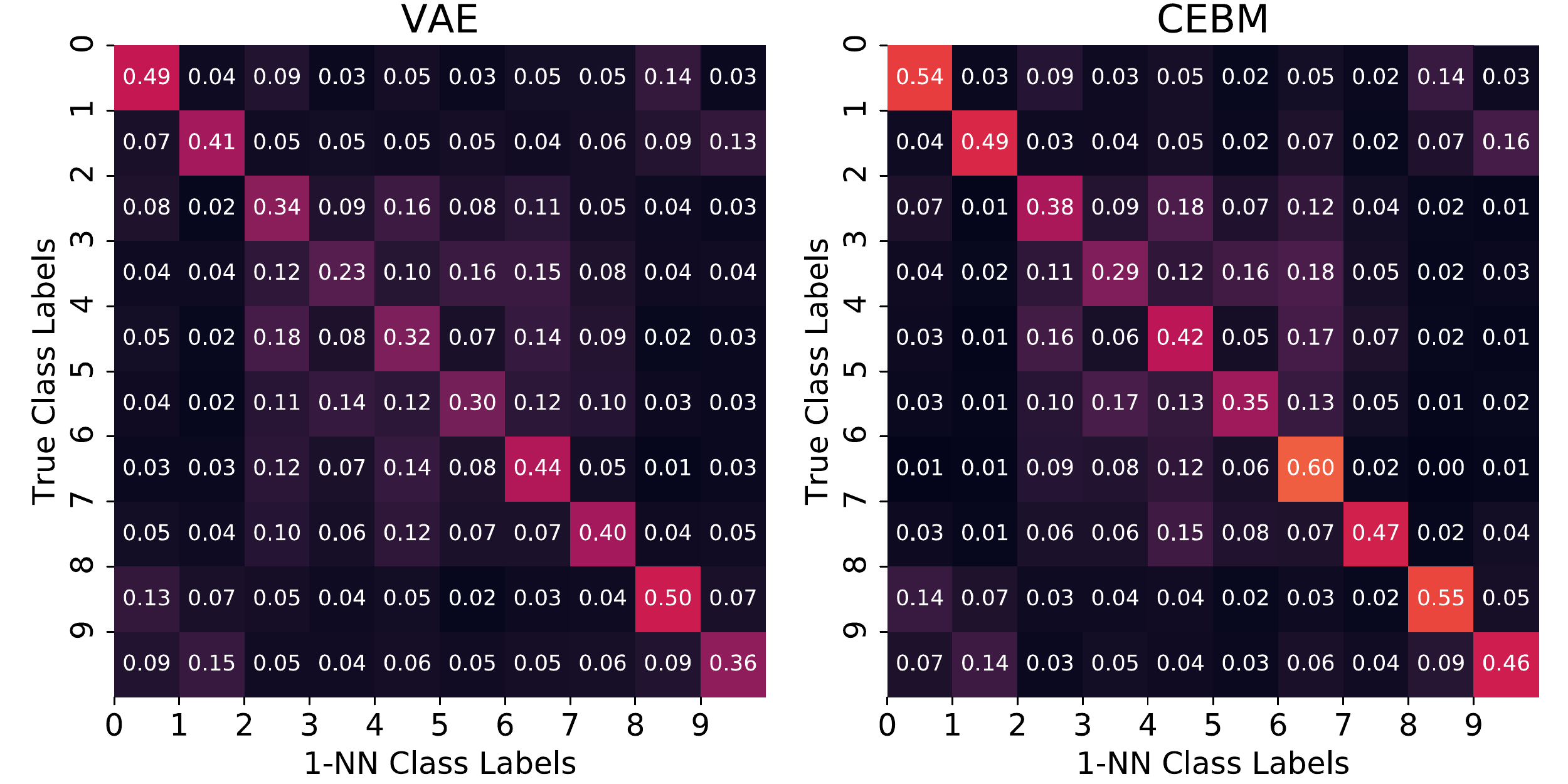}
\end{subfigure}
\vspace*{-1ex}
\caption{(\emph{Left}) Samples from CIFAR-10 along with the top 2-nearest-neighbors in pixel space, the latent space of a VAE, and the latent space of a CEBM. (\emph{Right}) Confusion matrices of 1-nearest-neighbor classification on CIFAR-10 based on L2 distance in the latent space. On average, CEBM representations more closely align with class labels compared to VAE.}
\label{fig:nearest-neighbours}
\vspace*{-1.5ex}
\end{figure*}

\section{Related Work}
\label{sec:related-work}

\vspace{-0.2\baselineskip}
\paragraph{Energy-Based Latent-Variable Models.} The idea of using EBMs to jointly model data and latent variables has a long history in the machine learning literature. Examples of this class of models include restricted Boltzmann machines (RBMs,~\citep{smolensky1986information,hinton2002training}), deep belief nets (DBNs,~\cite{hinton2006fast}), and deep Boltzmann machines (DBMs,~\citep{salakhutdinov2009deep}). 
The idea of extending RBMs in exponential families and exploiting conjugacy to yield a tractable posterior is also not new and has been explored in Exponential Family Harmoniums~(EFHs; \citep{welling2005exponential}). These models differ from CEBMs in that the they employ a bilinear interaction term $x^\top W z$, which ensures that both the likelihood $p(x \mid z)$ and $p(z \mid x)$ are tractable. In CEBMs, the corresponding term $t_\theta(x)^\top z$ is nonlinear, which means that the posterior is tractable, but the likelihood is not. 
We provide a more detailed discussion regarding the connection of our work to this class of models in Appendix~\ref{app:sec:rbm}.

\vspace{-0.2\baselineskip}
\paragraph{EBMs for Image Modelling.} Recent work has shown that EBMs with convolutional energy functions can accurately model distributions over images \citep{xie2016theory,nijkamp2019anatomy,nijkamp2019learning, du2019implicit,xie2021cooperative}. This line of work typically focuses on generation and not on unsupervised representation learning as we do here. A line of work, which is similar to ours in spirit, employs EBMs as priors on the latent space of deep generative models~\citep{pang2020learning,aneja2020ncp}. These approaches, unlike our work, require a generator.

\vspace{-0.2\baselineskip}
\paragraph{Interpretation of other models as EBMs.} \citet{grathwohl2019your,liu2020hybrid,xie2016theory} have proposed to interpret a classifier as an EBM that defines a joint energy function on the data and labels. CEBMs with a discrete bias can interpreted as the unsupervised variant of this model class.
\citet{che_your_2020} interpret a GAN as an EBM defined by both the generator and discriminator.

\vspace{-0.2\baselineskip}
\paragraph{Training EBMs.} A commonly used training method is PCD~\citep{tieleman2008training}, where the MCMC is initialized from a replay buffer that stores the previously generated samples~\citep{du2019implicit}, or from a generator~\citep{xie2018cooperative,xie_cooperative_2020,xie2021cooperative}. \citet{nijkamp2019anatomy,nijkamp2019learning} comprehensively investigate the convergence of PCD based on a variety of factors such as MCMC initialization, network architecture, and the optimizer. They find that the difference between the energy of the data and model samples is a good diagnostic of training stability. Many of these findings were helpful during the training and evaluation in our work.%

There is a large literature on alternative training methods. \citet{gao2020flow} propose to use the noise contrastive estimation (NCE,~\citep{gutmann2010noise}), where they pretrain a flow-based noise model and then train the EBM to discriminate between the real data examples and the ones generated from the noise model. 
Another popular approach is the score matching (SM,~\citet{vertes2016learning,hyvarinen2005estimation,vincent2011connection,song2020sliced,bao_bi-level_2020}), which learns EBMs by matching the gradient of the log probability density of the model distribution to that of the data distribution. \citet{bao_bi-level_2020} propose a bi-level version of this method where it is also applicable to latent-variable models. 
To sidestep the need or MCMC sampling, \citet{han2019divergence,Han_2020_CVPR,xie2021learning} jointly train an EBM with a VAE in an adversarial manner; \citet{grathwohl2021no} learn a generator by entropy regularization.
We refer the readers to~\citet{song2021train} for a more comprehensive discussion on training methods for EBMs.

\section{Experiments}
\label{sec:experiments}

Our experiments evaluate to what extent CEBMs can learn representations that encode meaningful factors of variation, whilst discarding details about the input that we would consider noise. This question is difficult to answer in generality, and in some sense not well-posed; whether a factor of variation should be considered signal or noise can depend on context. For this reason, our experiments primarily focus on the extent to which representations in CEBMs can recover the multimodal structure in datasets that are normally used for classification. %
While class labels are an imperfect proxy, in the sense that they do not reflect all factors of variation that we may want to encode in a representation, they provide a means of quantifying differences between representations that were learned in an unsupervised manner. 

We begin with a qualitative evaluation by visualizing samples and latent representation. We then demonstrate that learned representations align with class structure, in the sense that nearest neighbors in the latent space are more likely to belong to the same class (section~\ref{sec:exp:quality}). Next, we evaluate performance on out-of-distribution detection (OOD) tasks which, although not our primary focus in this paper, are a common use case for EBMs (Section~\ref{sec:exp:ood}).  %
We then quantify the extent to which the learned representations can improve performance in downstream task, we measure few-label classification accuracy for representations that were pre-trained without supervision (Section~\ref{sec:exp:fewshots}). Finally, we perform a more in-depth study of the latent space where we investigate to what extend the aggregate posterior distribution is close to the inductive bias as well how vulnerable CEBMs are to posterior collapse (Section~\ref{sec:exp:posterior-collapse}).

\setlength{\tabcolsep}{4.5pt}
\begin{table*}[!t]
\caption{AUROC scores in OOD Detection. We use $\log p_{\q}(\x)$ and $\| \nabla_{\x}\log p_{\q}(\x)\|$ as score functions.The left block shows results of the models trained on F-MNIST and tested on MNIST, E-MNIST, Constant (C); The right block shows results of the models trained on CIFAR-10 and tested on SVHN, Texture and Constant (C).}
\centering
\vspace*{0.5ex}
\begin{tabular}{l|ccc|ccc||ccc|ccc}
\toprule
& \multicolumn{6}{c||}{Fashion-MNIST} &    \multicolumn{6}{c}{CIFAR-10}\\
& \multicolumn{3}{c|}{$\log p_{\q}(\x)$} & \multicolumn{3}{c||}{$\| \nabla_{\x}\log p_{\q}(\x)\|$} &  \multicolumn{3}{c|}{$\log p_{\q}(\x)$} & \multicolumn{3}{c}{$\| \nabla_{\x}\log p_{\q}(\x)\|$}\\
\midrule
            &  MNIST  & E-MNIST& C &  MNIST & E-MNIST & C &  SVHN & Texture & C &  SVHN & Texture & C \\
\midrule
VAE         & .50 & .39 & .09 & .61 & .57 & .01  & .42 & \textbf{.58} & .41 & .38 & \textbf{.51} & .37 \\
IGEBM       & .35 & .36 & .90 & .78 & .82 & .96 & .45 & .31 & .64 & .33 & .17 & \textbf{.62} \\
CEBM        & .37 & .34 & .90 & \textbf{.82} & \textbf{.89} & \textbf{.98} & .47 & .32 & \textbf{.66} & .31 & .17 & .54 \\
GMM-CEBM       & \textbf{.56} & \textbf{.56} & \textbf{.92} & .56 & .80 & .95     & \textbf{.55} & .30 & .62 & \textbf{.40} & .23 & \textbf{.62}  \\
\bottomrule
\end{tabular}
\label{tab:ood-detection}
\vspace*{-1.0ex}
\end{table*}

\subsection{Network Architectures and Training}

\vspace{-0.2\baselineskip}
\paragraph{Architectures \& Optimization.} The CEBMs in our experiments employ an encoder network $t_\q(x)$ in the form of 4-layer CNN (as proposed by~\citet{nijkamp2019anatomy}), followed by an MLP output layer. We choose the dimension of latent variables to be 128. We found that the optimization becomes difficult with smaller dimensions. We train our models using 60 SGLD steps, 90k gradient steps, batch size 128, Adam optimizer with learning rate 1e-4. For training stability, we L2 regularize energy magnitudes (proposed by~\citet{du2019implicit}). See Appendix~\ref{app:sec:training-details} for details.

\setlength{\tabcolsep}{6pt}
\begin{table*}[!t]
\caption{Average classification accuracy on the test set. We train a variety of deep generative models on MNIST, Fashion-MNIST, CIFAR-10, and SVHN in an unsupervised way. Then we use the learned latent representations to train logistic classifiers with 1, 10, 100 training examples per class, and the full training set. We train each classifier 10 times on randomly drawn training examples.} 
\centering
\vspace*{0.5ex}
\begin{tabular}{l|cccc||cccc||cccc||cccc}
\toprule
 & \multicolumn{4}{c||}{MNIST} & \multicolumn{4}{c||}{Fashion-MNIST} & \multicolumn{4}{c||}{CIFAR-10} & \multicolumn{4}{c}{SVHN}\\
Models & $1$ & $10$  & $100$ & $full$ & $1$ & $10$  & $100$ & $full$ & $1$ & $10$  & $100$ & $full$ & $1$ & $10$  & $100$ & $full$ \\
\midrule
VAE & 42 & 85 & 92 & 95 & 41 & 63 & 72 & 81 & 16 & 22 & 31 & 38 & \textbf{13} & 13 & 16 & 36\\
GMM-VAE & 53 & 86 & 93 & 97 & 49 & 68 & 79 & 84 & \textbf{19} & 23 & 33 & 39 & \textbf{13} & 14 & 23 & 56  \\
BIGAN & 33 & 67 & 85 & 91 & 46 & 65 & 75 & 81 & 18 & \textbf{30} & \textbf{43} & 52 & 11 & 20 & 42 & 56  \\ 
\midrule
IGEBM & 63 & 89 & 95 & 97 & 50 & \textbf{70} & 79 & 83 & 16 & 26 & 33 & 42 & 10 & 16 & 35 & 49\\
CEBM & \textbf{67} & 89 & 95 & 97 & \textbf{52} & \textbf{70} & 77 & 83 & \textbf{19} & \textbf{30} & 42 & \textbf{53} & 12 & \textbf{25} & \textbf{48} & \textbf{70} \\
GMM-CEBM & \textbf{67} & \textbf{91} & \textbf{97} & \textbf{98} & \textbf{52} & \textbf{70} & \textbf{80} & \textbf{85} & 16 & 29 & 42 & 52 & 10 & 17 & 39 & 60 \\
\bottomrule
\end{tabular}
\label{tab:few-shot classification}
\end{table*}

\vspace{-0.5\baselineskip}
\paragraph{Hyperparameter Sensitivity.} As observed in previous work~\cite{du2019implicit,grathwohl2019your}, training EBMs is challenging and often requires a thorough hyperparameters search. We found that the choices of activation function, learning rate, number of SGLD steps, and regularization will all affect training stability. Models regularly diverge during training, and it is difficult to perform diagnostics given that $\log p_{\q,\lambda}(x)$ cannot be computed. As suggested by~\cite{nijkamp2019anatomy}, we found checking the difference in energy between data and model samples can help to verify training stability. In general we also observed a trade-off between sample quality and the predictive power of latent variables in our experiments. We leave investigation of the source of this trade-off to future work, but we suspect that this is because SGLD has more difficulty converging when the latent space is more disjoint.

\vspace*{-1.0ex}
\subsection{Samples and Latent Space}
\label{sec:exp:quality}
\vspace*{-1.0ex}

We begin with a qualitative evaluation by visualizing samples from the model. 
While generation is not our intended use case in this paper, such samples do serve as a diagnostic that allows us to visually inspect what characteristics of the input data are captured by the learned representation. 

Figure~\ref{fig:generated-samples} shows samples from CEBMs trained on MNIST, Fashion-MNIST, SVHN, and CIFAR-10. We initialize the samples with uniform noise and run 500 SGLD steps. We observe that the distribution over images is diverse and captures the main characteristics of the dataset. Sample quality is roughly on par with samples from other EBMs \cite{nijkamp2019anatomy}, although it is possible to generate samples with higher visual quality using class-conditional EBMs~\cite{du2019implicit, grathwohl2019your, liu2020hybrid} (which assume access to labels).

To assess to whether the representation in CEBMs aligns with classes in each dataset, we look at the agreement between the label of an input and that of its nearest neighbor in the latent space. The latent representations are inferred by computing the mean of the posterior $p_{\q,\lambda}(z|x)$. In Figure~\ref{fig:nearest-neighbours}, we show samples from CIFAR-10, along with the images that correspond to the nearest neighbors in pixel space, the latent space of a VAE, and the latent space of a CEBM. The distance in pixel space is a poor measure of similarity in this dataset, whereas proximity in the latent space is more likely to agree with class labels in both VAEs and CEBMs. We additionally show visualization of the latent space with UMAP~\citep{mcinnes2018umap} in Figure~\ref{app:fig:umap}.

In Figure~\ref{fig:nearest-neighbours} (right), we quantify this agreement by computing the fraction of neighbors in each class conditioned on the class of the original image. We see a stronger alignment between classes and the latent representation in CEBMs, which is reflected in higher numbers on the diagonal of the matrix. On average, a fraction of 0.38 of the nearest neighbors are in the same class in the VAE, whereas 0.45 of the neighbors are in the same class in the CEBM. This suggest that the representation in CEBMs should lead to higher performance in downstream classification tasks. We will evaluate this performance in Section~\ref{sec:exp:fewshots}.

\subsection{Out-of-Distribution Detection}\label{sec:exp:ood}
\vspace*{-1.0ex}

EBMs have formed the basis for encouraging results in out-of-distribution (OOD) detection~\cite{du2019implicit,grathwohl2019your}. While not our focus in this paper, OOD detection is a benchmark that helps evaluate whether a learned model accurately characterizes the data distribution. In Table~\ref{tab:ood-detection}, we report results in terms of two metrics. The first is the area under the receiver-operator curve (AUROC) when thresholding the log marginal $\log p_{\q,\lambda}(x)$. The second is the gradient-based score function proposed by ~\citet{grathwohl2019your}. We observe that in most cases, CEBM yields a similar score to the VAE and IGEBM baselines.

\subsection{Few-label Classification}\label{sec:exp:fewshots}
\vspace*{-1.0ex}

To evaluate performance in settings where few labels are available, we use pre-trained representations (which were learned without supervision) to train logistic classifiers with $1, 10, 100$ training examples per class, as well as the full training set. We evaluate classification performance for a spherical Gaussian bias (CEBM) and the mixture of Gaussians bias (GMM-CEBM). We compare our models against the IGEBM~\cite{du2019implicit} \footnote{Since the IGEBM does not explicitly have latent representations, we extract features from the last layer of the energy function.}, a standard VAE with the spherical Gaussian prior, GMM-VAE~\cite{tomczak2018vae} where the prior is a mixture of Gaussians (GMM), and BIGAN~\cite{donahue2016adversarial}.

We report the classification accuracy on the test set in Table~\ref{tab:few-shot classification}. CEBMs overall achieve a higher accuracy compared to VAEs in particular for CIFAR-10 and SVHN where the pixel distance is not good measure for similarity. Moreover, we observe that CEBMs outperform the IGEBM. This suggests that the inductive biases in CEBMs can lead to increased performance in downstream tasks. The performance between BIGANs and CEBMs is not as distinguishable which we suspect is due the fact BIGANs, just like CEBMs, do not define a likelihood that measure similarity at the pixel level. 
We also observe that the CEBM with the GMM inductive bias does not always outperform the one with the Gaussian inductive bias, which we suspect is due to GMM-CEBM having more difficulty to converge.

\subsection{Limitations: Posterior Collapse}
\label{sec:exp:posterior-collapse}
\vspace*{-0.25\baselineskip}
While our experiments demonstrate that CEBMs are able to reasonably approximate the data distribution and learn latent representations that are in closer agreement with class labels, they do not evaluate the learned notation of posterior uncertainty, and more generally the role of inductive bias. 
In this subsection, we ask the following two questions: (1) Does the aggregate posterior distribution of the training data live close to the inductive bias $p_\lambda(z)$? (2) What is the mutual information between latent variables and the training data?

\begin{table}[!t]

\vspace{-0.5\baselineskip}
\caption{KL divergence between aggregate posterior and prior and the mutual information between data and latent variables.} 
\vspace{0.5\baselineskip}
\begin{center}
\begin{tabular}{lcccccc}
\toprule
         & \multicolumn{2}{c}{VAE} & \multicolumn{2}{c}{CEBM} & \multicolumn{2}{c}{GMM-CEBM} \\
         & KL         & MI         & KL           & MI           & ~~~KL          & MI\\
\midrule
MNIST    &   11.5      & 9.1      & 0.9         & 0.3        & ~~~18.7           & 4.7 \\
FMNIST   &   3.5      & 9.0       & 0.6         & 0.4        & ~~~8.1           & 3.9 \\
CIFAR10  &   21.5      & 9.2      & 0.1         & 0.2        & ~~~4.5           & 2.7 \\
SVHN     &   8.6      & 10.1      & 0.1         & 0.1        & ~~~5.6           & 2.2 \\
\bottomrule
\end{tabular}
\vspace{-0.75\baselineskip}
\end{center}

\label{tab:kl-mi}
\end{table}

To evaluate whether encoded examples are distributed according to the bias $p_\lambda(z)$, we compute the divergence $\KL{\tilde{p}_{\q,\lambda}(z)}{p_{\lambda}(z)}$ between the bias and the aggregate posterior, which is a mixture over training data
\begin{align*}
    \hat{p}_{\q,\lambda}(z) = \frac{1}{N}\sum_n p_{\q,\lambda}(z|x_n), \quad x_n \sim p_{\text{data}}(x).
\end{align*}
There are two reasons to consider this distribution, rather than the marginal $p_{\theta\,\lambda}(z)$ of the CEBM. The first is computational expedience; it is easier to approximate $\hat{p}_{\q,\lambda}(z)$ than it is to approximate $p_{\q,\lambda}(z)$, since the latter requires samples $x \sim p_{\theta,\lambda}(x)$ from the marginal of the CEBM. The second reason is that $\hat{p}_{\q,\lambda}(z)$ reflects the distribution over features that we might use in a downstream task.

We approximate $\hat{p}_{\q,\lambda}(z)$ with a Monte Carlo estimate over batches of size 1k (see~\citet{esmaeili2019structured}), which we use to estimate both the KL and the mutual information (see Table~\ref{tab:kl-mi}). Because the marginal KL in CEBMs is significantly lower compared to VAEs across datasets,  we conclude that CEBMs indeed attempted to place the aggregate posterior distribution close to the inductive bias. 

Our evaluation of the mutual information proved more surprising: CEBMs learn a representation that has a very low mutual information between $x$ and $z$.
The reason for this is that the posterior parameters $\tilde{\lambda}_\theta(x) = \lambda + t_\theta(x)$ are dominated by the parameters of the bias $\lambda$, which means that model essentially ignores the sufficient statistics $t_\theta(x)$, which tend to have a small magnitude relative to $\lambda$. This phenomenon could be interpreted as an instance of \emph{posterior collapse}~\citep{alemi2017fixing}, which has been observed in a variety of contexts when training variational autoencoders by maximizing the marginal likelihood, which in itself is not an objective that guarantees a high mutual information.

\section{Discussion}
\label{sec:conclusion}

In this paper, we introduced CEBMs, a class of latent-variable models that factorize into an energy-based distribution over data and a tractable posterior over latent variables. CEBMs can be trained using standard methods for EBMs and in this sense have a small ``edit distance'' relative to existing approaches, whilst also providing a mechanism for incorporating inductive biases for latent variables.

Our experimental results are encouraging but also raise questions. We observe a closer agreement between the unsupervised representation and class labels than in VAEs, which translates into improvemed performance in downstream classification tasks. At the same time, we observe that CEBMs do not learn a meaningful notion of uncertainty; the CEBM posterior is typically dominated by the inductive bias, which means that there is a very low mutual information between data and latent variables. 

This work opens up a number of lines of future research. First and foremost, this work raises the question what objectives would be most suitable for learning energy-based latent-variable models in a manner maximizes agreement with respect to both the data distribution and the inductive bias terms, whilst also ensuring a sufficiently high mutual information between data and latent variables. More generally, we see opportunities to develop CEBMs with structured bias terms as an alternative to models based on VAEs in settings where we are hoping to reason about structured representations with little or no supervision. 

\section*{Acknowledgements}
\vspace*{-0.35\baselineskip}
We would like to thank our reviewers for their thoughtful comments, as well as Heiko Zimmermann, Will Grathwohl and Jacob Kelly for helpful discussions. This work was supported by the Intel Corporation, the 3M Corporation, NSF award 1835309, startup funds from Northeastern University, the Air Force Research Laboratory (AFRL), and DARPA.

\bibliography{references}
\bibliographystyle{icml2021}

\appendix

\newpage
\onecolumn
\icmltitle{Supplementary Materials for Conjugate Energy-Based Models}
\appendix

\section{Connection to Exponential Family Harmoniums}
\label{app:sec:rbm}

As mentioned in Section~\ref{sec:related-work}, there is a long history of incorporating latent variables in EBMs, particularly in the context of restricted Boltzmann machines (RBMs)~\citep{smolensky1986information,hinton2002training}, deep belief nets~\citep{hinton2006fast}, and deep Boltzmann machines~\citep{salakhutdinov2009deep}. Moreover, the idea of formulating EBMs into the exponential family is also not new; \citet{welling2005exponential} proposed a new class of models called Exponential Family Harmoniums (EFHs) by extending RBMs into the exponential family. In this Section, we discuss the connection between our approach to these models. Concretely, we show that EFHs can be recovered a special case of CEBMs.

For observed variable $x$ and latent variable $z$, the energy of an RBM is defined as
\begin{align}
    \label{eq:rbm}
    E^{\text{RBM}}_{\q}(x,z)
    = -\inner{x^\top\theta_{xz}}{z} - \inner{x}{\theta_x} - \inner{z}{\theta_z},
\end{align}
where $\theta_x \in \mathbb{R}^D$, $\theta_z \in \mathbb{R}^K$, and $\theta_{xz} \in \mathbb{R}^{D \times K}$. In RBMs, the conditional distributions $p_{\q}(x|z)$ and $p_{\q}(z|x)$ are both tractable which means that during contrastive divergence, we can sample $x \sim p_\q(x)$ using Gibbs sampling.

EFHs extend these models into the exponential family by incorporating the sufficient statistics of $x$ and $z$ in the energy,
\begin{align}
    \label{eq:efh}
    E^{\text{EFH}}_{\q}(x,z)
    = -\inner{t_x(x)^\top\theta_{xz}}{t_z(z)} - \inner{t_x(x)}{\theta_x} - \inner{t_z(z)}{\theta_z},
\end{align}
where $t_x(\cdot)$ and $t_z(\cdot)$ are the sufficient statistics for variables $x$ and $z$ respectively. \citet{welling2005exponential} show that this energy function yields the following conditional distributions:
\begin{align}
    \label{eq:efh-conditional}
    \text{Likelihood} \qquad p_{\q}(x|z) &= \exp \left\{ \inner{t_x(x)}{\tilde{\q}_x} - A(\tilde{\q}_x)\right\}, \qquad \tilde{\q}_x = \q_x + \theta_{xz}t_z(z), \\
    \text{Posterior} \qquad p_{\q}(z|x) &= \exp \left\{ \inner{t_z(z)}{\tilde{\q}_z} - B(\tilde{\q}_z)\right\}, \qquad \tilde{\q}_z = \q_z + \theta_{xz}t_x(x),
\end{align}
where $\tilde{\q}_x$ and $\tilde{\q}_z$ are the canonical parameters, and $A(\cdot)$ and $B(\cdot)$ are the log normalizer of the models $p_{\q}(x|z)$ and $p_{\q}(z|x)$ respectively. Given that both conditional distributions are tractable, EFHs have the same advantage as RBMs: We can use a Gibbs sampler for sampling $x \sim p_\q(x)$.

CEBMs can be considered an extension of EFHs. In Equation~\ref{eq:cebm-energy}, we recover the energy function for an EFH by setting
\begin{align}
    t_{\q}(x)=[t_x(x)^\top\theta_{xz},\ \inner{\theta_{x}}{t_x(x)}], \qquad \eta(z)=[t_z(z),\ 1], \qquad E_{\q}(z)=-\inner{t_z(z)}{\theta_{z}}.
\end{align}
Perhaps the most crucial difference between CEBMs and EFHs (and other RBM-based models) is the non-linearity relationship between the observed and latent variables. The non-linearity in $t_{\q}(\cdot)$ has the benefit of providing the flexibility to learn more complex structures in the data. This modelling choice however comes with a cost. In CEBMs, while the posterior is still tractable, the likelihood model is not. As a consequence, we lose the ability to use Gibbs sampling to sample $x \sim p_\q(x)$. However, given that our motivation here is not to generate high quality samples at test time but to learn good representations representations, we believe giving up the ability to easily sample $x$ in order to learn more complex structures while keeping the posterior tractable is an appropriate trade-off.

\section{Derivation of Prior and Likelihood in a CEBM}
\label{app:sec:derivations}

\renewcommand{\arraystretch}{1.1}
\begin{table}[!t]
\centering
\begin{tabular}{lll}
    \toprule
    \textbf{Energy Type} & \textbf{Model} & \textbf{Energy} \\
    \midrule
    $E_\q(x)$ & IGEBM~\citep{du2019implicit} & $f_\q(x)$\\
    \midrule
    $E_\q(x,y)$ & JEM~\citep{grathwohl2019your} & $-f_\q(x)[y]$\\
                & HDGE~\citep{liu2020hybrid} & \\
    \midrule
    $E_\q(x,z)$ & RBM~\citep{smolensky1986information} & $-\inner{x^\top\q_{x,z}}{z} - \inner{x}{\theta_x} - \inner{z}{\theta_z}$\\
                & EFH~\citep{welling2005exponential} & $-\inner{t(x)^\top\q_{x,z}}{t(z)} - \inner{t(x)}{\theta_x} - \inner{t(z)}{\theta_z}$ \\
                & VAE~\citep{kingma2013auto-encoding} & $-\inner{x}{\mu_\q(z)} + A(\eta_\q(z)) + E(z) $\\
                & GAN~\citep{che_your_2020} & $D_\q(x) + E(z)$\\
                & CEBM~(this paper) & $-\inner{t_\q(x)}{\eta(z)} + E(z)$\\
    \midrule
    $E_\q(x,y,z)$   & GMM-VAE~\citep{tomczak2018vae} & $-\inner{x}{\mu_\q(z,y)} + A(\eta_\q(z,y)) + E(z,y) $ \\ 
                & GMM-CEBM~(this paper) & $-\inner{t_\q(x)}{\eta(y,z)} + E(y,z)$\\
    \bottomrule
\end{tabular}
\caption{Comparison of energies in generative models. The functions $f_\q(\cdot)$, $\eta_\q(\cdot)$, and $t_\q(\cdot)$ are typically deep neural networks (DNNs). In EBMs defined on only the data space (type $E_\q(x)$) such as IGEBM, the DNN outputs a scalar value $f_\q(x):\mathbb{R}^D \rightarrow \mathbb{R}$. In EBMs defined on the data space as well as labels (type $E_\q(x,y)$) such as JEM, the DNN outputs a vector of length $L$ corresponding to the number of classes $f_\q(x):\mathbb{R}^D \rightarrow \mathbb{R}^L$. In GAN, $D_\q(x)$ refers to the discriminator. }
\label{tab:energy-table}
\end{table}
\renewcommand{\arraystretch}{1.0}
\subsection{Prior}

\begin{align}
    p_{\q,\lambda}(z) 
    &=  \int \!dx \: \frac{1}{Z_{\q,\lambda}}\exp \{- E_{\q,\lambda}(x,z)\} \\
    &= \frac{1}{Z_{\q,\lambda}} \int \!dx \: \exp \{- E_{\q,\lambda}(x,z)\} \\
    &= \frac{1}{Z_{\q,\lambda}} \int \!dx \: \exp \{\inner{t_\q(x)}{\eta(z)} - E_\lambda(z)\} \\
    &= \frac{\exp\{-E_\lambda(z)\}}{Z_{\q,\lambda}} \int \!dx \: \exp \{\inner{t_\q(x)}{\eta(z)}\}
\end{align}

\subsection{Likelihood}

\begin{align}
    p_{\q,\lambda}(x|z) 
    &= \frac{p_{\q,\lambda}(x,z)}{p_{\q,\lambda}(z)} \\
    &= \frac{\frac{1}{Z_{\q,\lambda}}\exp \{- E_{\q,\lambda}(x,z)\} }{\frac{\exp\{-E_\lambda(z)\}}{Z_{\q,\lambda}} \int \!dx \: \exp \{\inner{t_\q(x)}{\eta(z)}\}} \\
    &= \frac{\frac{\exp\{-E_\lambda(z)\}}{Z_{\q,\lambda}} \exp \{\inner{t_\q(x)}{\eta(z)}\} }{\frac{\exp\{-E_\lambda(z)\}}{Z_{\q,\lambda}} \int \!dx \: \exp \{\inner{t_\q(x)}{\eta(z)}\}} \\
    &= \frac{\exp \{\inner{t_\q(x)}{\eta(z)}\} }{\int \!dx \: \exp \{\inner{t_\q(x)}{\eta(z)}\}}
\end{align}

\section{Training Details}
\label{app:sec:training-details}
In CEBMs and VAEs, we choose the dimension of latent variables to be 128. For CEBMS, We found that the optimization becomes difficult with smaller dimensions. We L2 regularize energy magnitudes (proposed by~\citet{du2019implicit}), where the coefficient of the L2 regularization term is 0.1. We empirically found that the training would become unstable without this regularization. We train our models using 60 SGLD steps where we initialize samples from the replay buffer with 0.95 probability, and initialize from uniform noise with 0.05 probability. We train all the models with 90k gradient steps, batch size 128, Adam optimizer with learning rate 1e-4. When doing PCD, we used a reply buffer of size 5000. We set the $\alpha$ in the SGLD teps to be 0.075. Similar to~\citet{du2019implicit}, we found it useful to add some noise to the image before encoding. In our experiments, we used Gaussian noise with $\sigma^{2} = 0.03$. We used 50 GMM components for GMM-VAE and 10 GMM components for GMM-CEBM.  

\newpage
\vspace*{-2ex}
\section{Model Architectures}
\label{appendix-architectures}
\vspace*{-1ex}
Table~\ref{app:tab:arch-cebm}, Table~\ref{app:tab:arch-vae}, and Table~\ref{app:tab:arch-igebm} show the architectures used for CEBM, VAE, and IGEBM, respectively.

\begin{table}[!h]
\vspace*{-3ex}
\caption{Architecture of CEBM and GMM-CEBM}
    \centering
    \begin{subtable}[h]{.5\linewidth}
    \caption{MNIST and Fashion-MNIST.}
    \centering
        \begin{tabular}{|l|}
        \toprule
        \textbf{Encoder} \\
        \midrule
        Input $28\times28\times1$ images  \\
        \hline 
        $3\times3$ conv. 64 stride 1. padding 1. Swish.  \\
        \hline 
        $4\times4$ conv. 64 stride 2. padding 1. Swish. \\
        \hline 
        $4\times4$ conv. 32 stride 2. padding 1. Swish.  \\
        \hline
        $4\times4$ conv. 32 stride 2. padding 1. Swish. \\
        \hline
        FC. 128 Swish. \\
        \hline
        FC. $2\times128$ \\
        \bottomrule
        \end{tabular}
    \end{subtable}%
    \begin{subtable}[h]{.5\textwidth}
    \caption{CIFAR10 and SVHN.}
    \centering
        \begin{tabular}{|l|}
        \toprule
        \textbf{Encoder}  \\
        \midrule
        Input $32\times32\times3$ images  \\
        \hline 
        $3\times3$ conv. 64 \hspace{0.2em} stride 1. padding 1. Swish.  \\
        \hline 
        $4\times4$ conv. 128 stride 2. padding 1. Swish. \\
        \hline 
        $4\times4$ conv. 256  stride 2. padding 1. Swish.  \\
        \hline
        $4\times4$ conv. 512  stride 2. padding 1. Swish.  \\
        \hline
        FC. 1024 Swish. \\
        \hline
        FC. $2\times128$\\
        \bottomrule
        \end{tabular}
    \end{subtable}
    \label{app:tab:arch-cebm}
\end{table}

\begin{table}[!h]
\vspace*{-5ex}
\caption{Architecture of IGEBM}
    \centering
    \begin{subtable}[h]{.5\textwidth}
    \caption{MNIST and Fashion-MNIST.}
    \centering
        \begin{tabular}{|l|}
        \toprule
        \textbf{Encoder} \\
        \midrule
        Input $28\times28\times1$ images  \\
        \hline 
        $3\times3$ conv. 64  stride 1. padding 1. Swish.  \\
        \hline 
        $4\times4$ conv. 64  stride 2. padding 1. Swish. \\
        \hline 
        $4\times4$ conv. 32  stride 2. padding 1. Swish.  \\
        \hline
        $4\times4$ conv. 32  stride 2. padding 1. Swish. \\
        \hline
        FC. 128 Swish. \\
        \hline
        FC. 128 Swish. FC. 1 \\
        \bottomrule
        \end{tabular}
    \end{subtable}%
    \begin{subtable}[h]{.5\textwidth}
    \caption{CIFAR10 and SVHN.}
    \centering
        \begin{tabular}{|l|}
        \toprule
        \textbf{Encoder}  \\
        \midrule
        Input $32\times32\times3$ images  \\
        \hline 
        $3\times3$ conv. 64 \hspace{0.2em} stride 1. padding 1. Swish.  \\
        \hline 
        $4\times4$ conv. 128  stride 2. padding 1. Swish. \\
        \hline 
        $4\times4$ conv. 256 stride 2. padding 1. Swish.  \\
        \hline
        $4\times4$ conv. 512  stride 2. padding 1. Swish. \\
        \hline
        FC. 1024 Swish \\
        \hline
        FC. 128 Swish. FC. 1\\
        \bottomrule
        \end{tabular}
    \end{subtable}
    \label{app:tab:arch-igebm}
\end{table}

\vspace*{-5ex}
\begin{table}[!h]
\caption{Architecture of VAE and GMM-VAE}
    \centering
        \vspace*{1ex}
    \begin{subtable}[h]{\textwidth}
    \caption{MNIST and Fashion-MNIST.}
    \centering
        \begin{tabular}{|l|l|}
        \toprule
        \textbf{Encoder} & \textbf{Decoder} \\
        \midrule
        Input $28\times28\times1$ images & Input $z\in \mathbb{R}^{128}$ latent variables \\
        \hline 
        $3\times3$ conv. 64 stride 1. padding 1. ReLU. & FC. 128 ReLU. FC. $3\times3\times32$ ReLU. \\
        \hline 
        $4\times4$ conv. 64 stride 2. padding 1. ReLU. & $4\times4$ upconv. 32 stride 2. padding 1. ReLU. \\
        \hline 
        $4\times4$ conv. 32 stride 2. padding 1. ReLU. & $4\times4$ upconv. 64 stride 2. padding 1. ReLU. \\
        \hline
        $4\times4$ conv. 32 stride 2. padding 1. ReLU. & $4\times4$ upconv. 64 stride 2. padding 0. ReLU. \\
        \hline
        FC. 128 ReLU. FC. $2\times128$. & $3\times3$ upconv. 1 \hspace{0.2em} stride 1. padding 0 \\
        \bottomrule
        \end{tabular}
    \vspace*{1ex}
    \end{subtable}
    \begin{subtable}[h]{\textwidth}
    \caption{CIFAR10 and SVHN.}
    \centering
        \begin{tabular}{|l|l|}
        \toprule
        \textbf{Encoder} & \textbf{Decoder} \\
        \midrule
        Input $32\times32\times3$ images & Input $z\in \mathbb{R}^{128}$ latent variables \\
        \hline 
        $3\times3$ conv. 64 stride 1. padding 1. ReLU. & FC. 128 ReLU. FC. $4\times4\times512$ ReLU. \\
        \hline 
        $4\times4$ conv. 128  stride 2. padding 1. ReLU. &  $4\times4$ upconv. 32 stride 2. padding 1. ReLU. \\
        \hline 
        $4\times4$ conv. 256  stride 2. padding 1. ReLU. & $4\times4$ upconv. 64 stride 2. padding 1. ReLU. \\
        \hline
        $4\times4$ conv. 512  stride 2. padding 1. ReLU. & $3\times3$ upconv. 64 stride 2. padding 1. ReLU. \\
        \hline
        FC. 1024 ReLU. FC. $2\times128$. & $3\times3$ upconv. 1 \hspace{0.2em} stride 1. padding 1 \\
        \bottomrule
        \end{tabular}
    \vspace*{1ex}
    \end{subtable}
    \label{app:tab:arch-vae}
    \vspace*{5ex}
\end{table}

\begin{table}[!t]
\caption{Architecture of BIGAN for MNIST and Fashion-MNIST.}
    \centering
        \vspace*{1ex}
    \begin{subtable}[h]{\textwidth}
    \caption{MNIST and Fashion-MNIST.}
    \centering
        \begin{tabular}{|l|}
        \toprule
        \textbf{Discriminator}  \\
        \midrule
        Input $28\times28\times1$ images  \\
        \hline 
        $3\times3$ conv. 64 stride 1. padding 1. BN. LeakyReLU. \\
        \hline 
        $4\times4$ conv. 64 stride 2. padding 1. BN. LeakyReLU. \\
        \hline 
        $4\times4$ conv. 32 stride 2. padding 1. BN. LeakyReLU. \\
        \hline
        $4\times4$ conv. 32  stride 2. padding 1. BN. LeakyReLU.  \\
        \hline
        FC. 128 LeakyReLU.   \\
        \hline
        256. FC 128 LeakyReLU. FC. $1$. Sigmoid. \\
        \bottomrule
        \end{tabular}
    \vspace*{1ex}
    \end{subtable}
    \begin{subtable}[h]{\textwidth}
    \centering
        \begin{tabular}{|l|l|}
        \toprule
        \textbf{Generator} & \textbf{Encoder} \\
        \midrule
         Input $z\in \mathbb{R}^{128}$ latent variables & Input $28\times28\times1$ images \\
        \hline 
        $4\times4$ upconv. 64 stride 1. padding 1. BN. ReLU. & $3\times3$ conv. 64 stride 1. padding 1. BN. LeakyReLU. \\
        \hline 
        $4\times4$ upconv. 64 stride 2. padding 1. BN. ReLU. & $4\times4$ conv. 64 stride 2. padding 1. BN. LeakyReLU.\\
        \hline
        $3\times3$ upconv. 32 stride 2. padding 1. BN. ReLU. & $4\times4$ conv. 32 stride 2. padding 1. BN. LeakyReLU.\\
        \hline
        $4\times4$ upconv. 32 stride 2. padding 1. BN. ReLU. & $4\times4$ conv. 32  stride 2. padding 1. BN. LeakyReLU.\\
        \hline
        $4\times4$ upconv. 1 \hspace{0.2em} stride 2. padding 1. Tanh. & FC. 128 LeakyReLU. FC. $2\times128$. \\
        \bottomrule
        \end{tabular}
    \vspace*{1ex}
    \end{subtable}
    \begin{subtable}[h]{\textwidth}
    \caption{CIFAR10 and SVHN.}
    \centering
        \begin{tabular}{|l|}
        \toprule
        \textbf{Discriminator}  \\
        \midrule
        Input $28\times28\times1$ images  \\
        \hline 
        $3\times3$ conv. 64 stride 1. padding 1. BN. LeakyReLU. \\
        \hline 
        $4\times4$ conv. 128 stride 2. padding 1. BN. LeakyReLU. \\
        \hline 
        $4\times4$ conv. 256 stride 2. padding 1. BN. LeakyReLU. \\
        \hline
        $4\times4$ conv. 512  stride 2. padding 1. BN. LeakyReLU.  \\
        \hline
        FC. 128 LeakyReLU. \\
        \hline
        256 FC 128 LeakyReLU. FC. $1$. Sigmoid.  \\
        \bottomrule
        \end{tabular}
    \vspace*{1ex}
    \end{subtable}
    \begin{subtable}[h]{\textwidth}
    \centering
        \begin{tabular}{|l|l|}
        \toprule
        \textbf{Generator} & \textbf{Encoder}\\
        \midrule
         Input $z\in \mathbb{R}^{128}$ latent variables & Input $28\times28\times1$ images \\
        \hline 
        $4\times4$ upconv. 512 stride 2. padding 1. BN. ReLU. & $3\times3$ conv. 64 stride 1. padding 1. BN. LeakyReLU.\\
        \hline 
        $4\times4$ upconv. 256 stride 2. padding 1. BN. ReLU. & $4\times4$ conv. 128 stride 2. padding 1. BN. LeakyReLU.\\
        \hline
        $4\times4$ upconv. 128 stride 2. padding 1. BN. ReLU. &  $4\times4$ conv. 256 stride 2. padding 1. BN. LeakyReLU.\\
        \hline
        $4\times4$ upconv. 64 stride 2. padding 1. BN. ReLU. & $4\times4$ conv. 512  stride 2. padding 1. BN. LeakyReLU.\\
        \hline
        $4\times4$ upconv. 3 \hspace{0.2em} stride 2. padding 1. Tanh. & FC. 128 LeakyReLU. FC $2\times128$.\\
        \bottomrule
        \end{tabular}
    \vspace*{1ex}
    \end{subtable}
    \label{appendix-tab:arch-bigan}
\end{table}

\newpage
\newpage

\section{Additional Results}
\label{app:sec:additional-results}

\subsection{Confusion Matrices on 1-NN Classification}
\label{appendix-sec:confuion matrices}
We perform 1-nearest-neighbor classification task for MNIST, Fashion-MNIST, SVHN, CIFAR10. We compute the L2 distance in the latent space of VAE, IGEBM and CEBM, and also in pixel space. We visualize the confusion matrices
\begin{figure}[!h]
\centering
\vspace*{-2ex}
    \begin{subfigure}[h]{\linewidth}
    \centering
    \includegraphics[width=\linewidth]{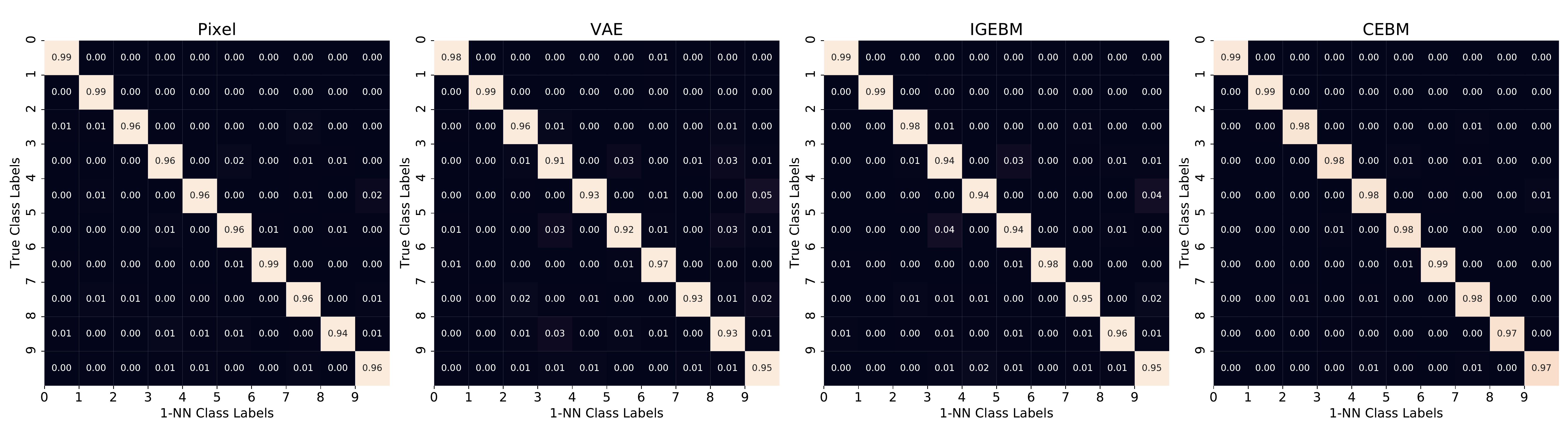}
    \caption{MNIST}
    \end{subfigure}
    \begin{subfigure}[h]{\linewidth}
    \centering
    \includegraphics[width=\linewidth]{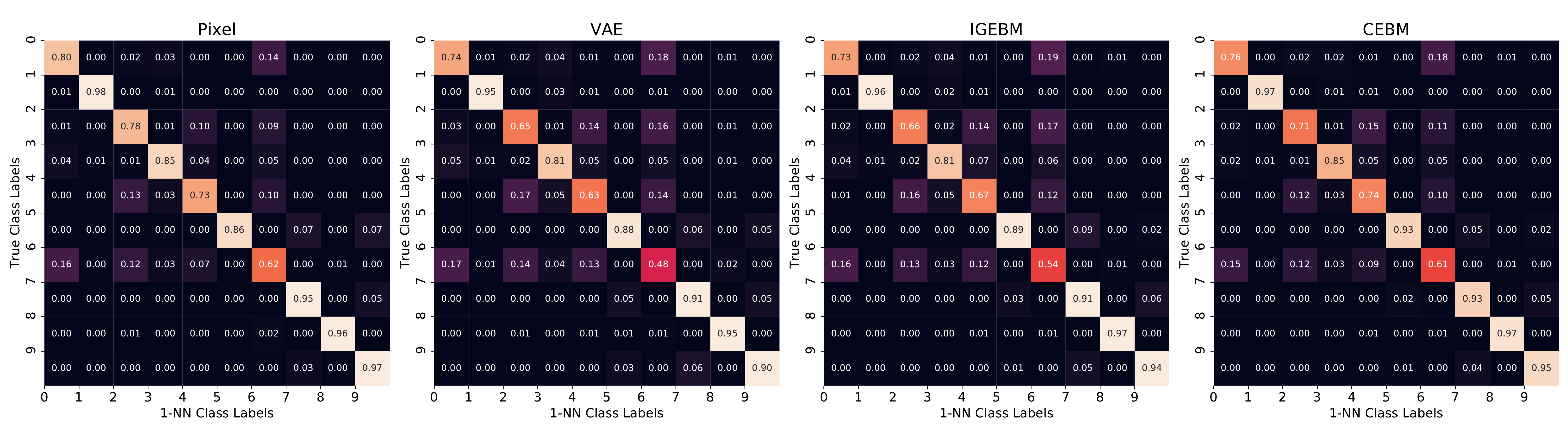}
    \caption{Fashion-MNIST}
    \end{subfigure}
    \begin{subfigure}[h]{\linewidth}
    \centering
    \includegraphics[width=\linewidth]{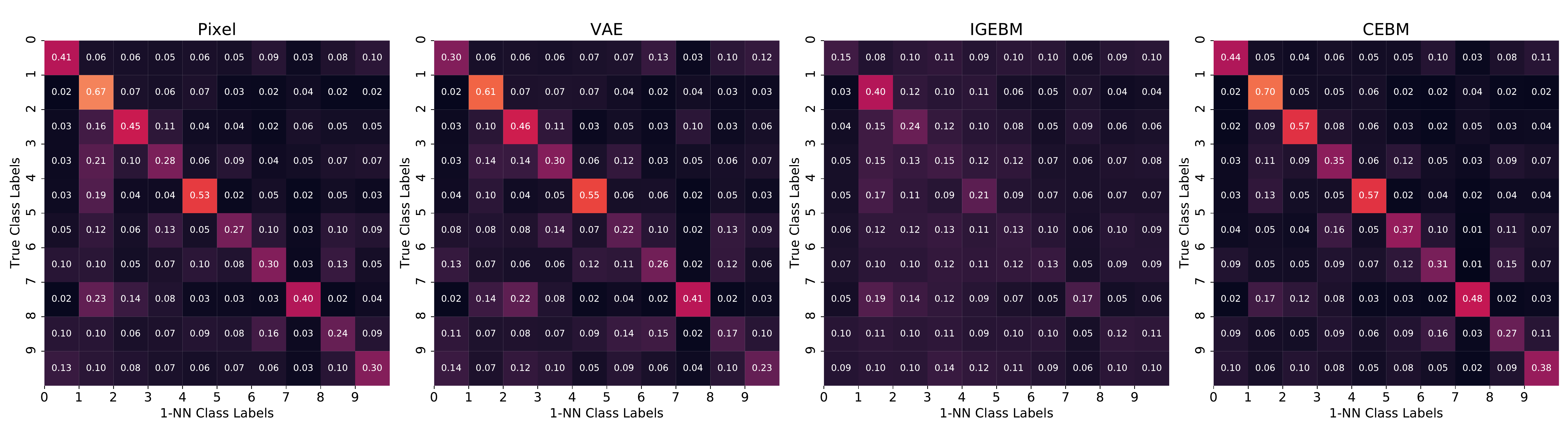}
    \caption{SVHN}
    \end{subfigure}
    \begin{subfigure}[h]{\linewidth}
    \centering
    \includegraphics[width=\linewidth]{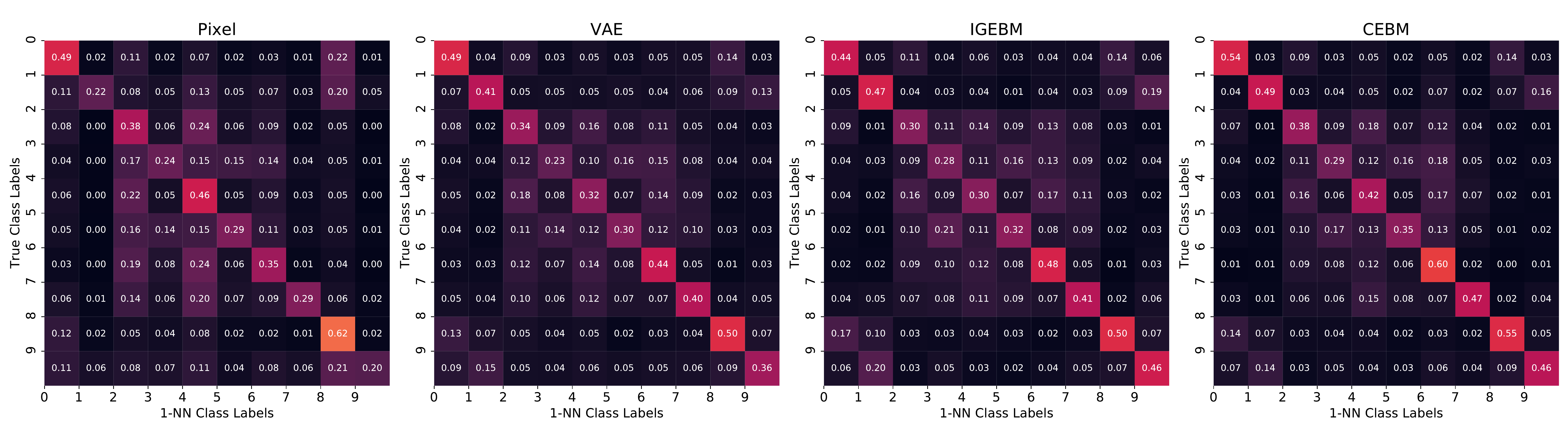}
    \caption{CIFAR10}
    \end{subfigure}
\label{appendix:confusion-matrices-mnist}
\end{figure}

\begin{figure}[!t]
    \centering
    \includegraphics[width=\textwidth]{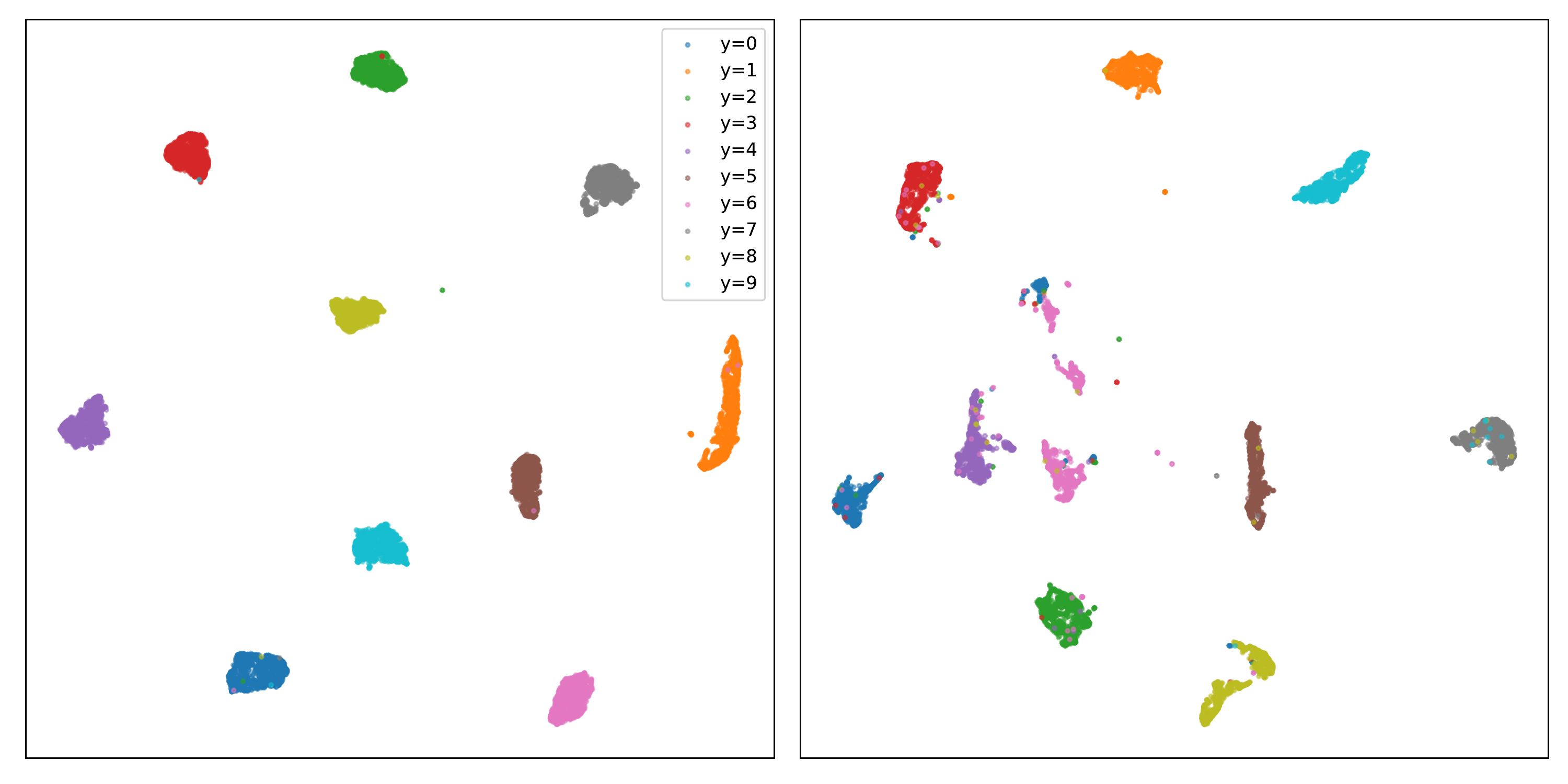}
    \caption{CEBMs latent space visualized with UMAP for MNIST (\emph{Left}) and FashionMNIST (\emph{Right}).}
    \label{app:fig:umap}
\end{figure}

\end{document}